\documentclass{article}
\usepackage{graphicx} 
\usepackage[margin=2.5cm]{geometry}
\usepackage{natbib}
\usepackage{amsmath, bm, amssymb}
\usepackage{mathrsfs}
\usepackage{hyperref}
\usepackage{booktabs}
\usepackage{multirow}
\usepackage{dsfont}
\usepackage{color}
\usepackage{enumitem}
\usepackage{indentfirst}
\usepackage{pifont}

\usepackage{amsthm}
\newtheorem{definition}{Definition}
\newtheorem{lemma}{Lemma}
\newtheorem{theorem}{Theorem}

\newtheorem{proposition}{Proposition}
\newtheorem{corollary}{Corollary}

\usepackage{apptools}
\AtAppendix{\counterwithin{lemma}{section}}
\AtAppendix{\counterwithin{corollary}{section}}
\AtAppendix{\counterwithin{remark}{section}}
\AtAppendix{\counterwithin{theorem}{section}}
\AtAppendix{\counterwithin{equation}{section}}

\hypersetup{hidelinks}

\title{\textbf{On Conditional Stochastic Interpolation for Generative Nonlinear Sufficient Dimension Reduction}}

\author{Shuntuo Xu$^1$, Zhou Yu$^1$, Jian Huang$^2$ \\
        oaksword98@gmail.com, zyu@stat.ecnu.edu.cn, j.huang@polyu.edu.hk \\
        $^1$School of Statistics, East China Normal University \\
        $^2$Departments of Data Science and Artificial Intelligence, and Applied Mathematics, \\
        The Hong Kong Polytechnic University}
\date{\today}

\begin{document}

\maketitle

\begin{abstract}
Identifying low-dimensional sufficient structures in nonlinear sufficient dimension reduction (SDR) has long been a fundamental yet challenging problem. Most existing methods lack theoretical guarantees of exhaustiveness in identifying lower dimensional structures, either at the population level or at the sample level. We tackle this issue by proposing a new method, generative sufficient dimension reduction (GenSDR), which leverages modern generative models. We show that GenSDR is able to fully recover the information contained in the central $\sigma$-field at both the population and sample levels. In particular, at the sample level, we establish a consistency property for the GenSDR estimator from the perspective of conditional distributions, capitalizing on the distributional learning capabilities of deep generative models. Moreover, by incorporating an ensemble technique, we extend GenSDR to accommodate scenarios with non-Euclidean responses, thereby substantially broadening its applicability. Extensive numerical results demonstrate the outstanding empirical performance of GenSDR and highlight its strong potential for addressing a wide range of complex, real-world tasks.
\end{abstract}

\section{Introduction}

Sufficient dimension reduction (SDR) is a powerful framework for supervised data compression that preserves all information in the covariates relevant to the response. It seeks a low-dimensional sufficient representation, typically a (possibly nonlinear) transformation of the original covariates, with applications in data visualization, improved classification, functional data analysis, and longitudinal association modeling \citep{li2018sufficient, meng2020sufficient, lee2022functional, Nghiem18022025}. Depending on the transformation class, SDR methods can be linear \citep{li1991sliced} or nonlinear \citep{lee2013general}. In this paper, we focus on the general nonlinear setting.

Formally, let $X \in \mathbb{R}^{d_x}$ denote the covariates and $Y \in \mathbb{R}^{d_y}$ the response. A sub-$\sigma$-field $\mathcal{G} \subseteq \sigma(X)$ is called an SDR $\sigma$-field \citep{lee2013general} if
\begin{equation}\label{eqn: nonlinear_sdr_sigma_field}
    Y \perp\!\!\!\perp X | \mathcal{G}.
\end{equation}
Thus, $\mathcal{G}$ captures all information in $X$ that is relevant for predicting $Y$. However, such a $\sigma$-field is not unique (for instance, $\sigma(X)$ itself always satisfies \eqref{eqn: nonlinear_sdr_sigma_field}). Under mild conditions, there exists a unique minimal $\sigma$-field $\mathcal{G}_{Y|X}$ that satisfies \eqref{eqn: nonlinear_sdr_sigma_field} and is contained in every SDR $\sigma$-field \citep{lee2013general}. This $\mathcal{G}_{Y|X}$ is referred to as the central $\sigma$-field. Since $\mathcal{G}_{Y|X}$ is abstract and difficult to estimate directly, it is typically assumed that $\mathcal{G}_{Y|X}$ is generated by a lower-dimensional nonlinear transformation $R_0$ of $X$, such that,
\begin{equation}\label{eqn: nonlinear_sdr_representation}
    Y \perp\!\!\!\perp X | R_0(X),
\end{equation}
where $R_0 : \mathbb{R}^{d_x} \to \mathbb{R}^d$ with $d \le d_x$. The goal of nonlinear SDR is to construct an estimator of a transformation equivalent to $R_0$, noting that $R_0$ itself is identifiable only up to $\mathcal{G}_{Y|X}$-measurability. A candidate transformation $R$ is called unbiased if $R(X)$ is $\mathcal{G}_{Y|X}$-measurable, and exhaustive if $\mathcal{G}_{Y|X} \subseteq \sigma(R(X))$. Unbiasedness and exhaustiveness respectively capture the accuracy and completeness of an SDR estimator.

Following the pioneering work of generalized sliced inverse regression (GSIR; \cite{lee2013general}) and generalized sliced average variance estimator (GSAVE; \cite{lee2013general}), a few methods have been proposed to estimate $R_0$. A persistent challenge, however, is establishing rigorous theoretical guarantees at both the population and sample levels, particularly on exhaustiveness. For instance, GSIR may fail to be unbiased and exhaustive at the population level when the central class, the set of all $\mathcal{G}_{Y|X}$-measurable and square-integrable functions, is incomplete; GSAVE is shown to be only unbiased under certain conditions. Generalized martingale difference divergence net (GMDDNet; \cite{chen2024deep}) does not address exhaustiveness, partly due to its close connection with inverse regression and GSIR. Deep dimension reduction (DDR; \cite{huang2024deep}) method demonstrates that $R_0$, when projected onto a Gaussian random vector, minimizes the population loss, but does not establish the unbiasedness or exhaustiveness of a general minimizer. Among recent developments, the belted and ensembled neural network (BENN; \cite{tang2024belted}) method arguably provides the strongest population-level exhaustiveness guarantees. BENN tackles nonlinear SDR via the conditional mean $\sigma$-field combined with an ensemble framework \citep{yin2011sufficient}, where a sufficiently rich ensemble class ensures that any eventual minimizer captures all information in $\mathcal{G}_{Y|X}$. Nevertheless, an analogous exhaustiveness result at the sample level is not available.

In this paper, we instead focus on modeling the conditional distribution of $Y$ given $X$, which fully characterizes the dependence of $Y$ on $X$. A related perspective was adopted by \cite{xia2007constructive} to estimate the central subspace in linear SDR. Assuming the existence of conditional densities, \cite{xia2007constructive} exploited the fact that the conditional density depends on $X$ only through a sufficient representation, and used kernel methods for nonparametric estimation. However, kernel-based approaches suffer from the curse of dimensionality and bandwidth selection issues \citep{gyorfi2002distribution, verleysen2005curse}, limiting their applicability in high-dimensional settings. While conditional distribution modeling is inherently challenging, recent advances in deep generative learning offer new tools. Modern generative models, including diffusion models \citep{ho2020denoising, yang2023diffusion} and flow-matching models \citep{liu2022flow, lipman2023flow}, not only achieve state-of-the-art generative performance but also enjoy strong theoretical foundations \citep{oko2023diffusion, fukumizu2025flow}. We adopt the conditional stochastic interpolation (CSI) framework \citep{albergo2023building, huang2023conditional} for its flexibility and generality.

In CSI models, conditional distribution learning is governed by conditional velocity fields. Analogous to the insight in \cite{xia2007constructive}, we show that, under mild conditions, the conditional velocity field associated with the sufficient structure in \eqref{eqn: nonlinear_sdr_representation} depends on $x$ only through $R_0(x)$. This nested structure enables direct estimation of $R_0$ from the conditional velocity field, which is often more tractable than the original conditional density. Furthermore, our framework can be extended to accommodate non-Euclidean responses. For responses taking values in general metric space, we project the raw responses into a Euclidean space via an ensemble class potentially induced by a cc-universal kernel \citep{zhang2024dimension}, and then construct an estimator of $R_0$ by aggregating parallel stochastic interpolation trajectories.

\begin{table}[t]
    \centering
    \small
    \caption{Comparison of features exhibited by existing methods and ours.}
    \begin{tabular}{ccccc}
        \toprule
        Method & Generative & Exhaustive & Distributional guarantee & Support non-Euclidean responses \\ \hline
        GSIR \citep{lee2013general} & \ding{55} & Limited & \ding{55} & \ding{55} \\
        GSAVE \citep{lee2013general} & \ding{55} & Limited & \ding{55} & \ding{55} \\
        DDR \citep{huang2024deep} & \ding{55} & \ding{55} & \ding{55} & \ding{55} \\
        GMDDNet \citep{chen2024deep} & \ding{55} & \ding{55} & \ding{55} & Potential \\
        BENN \citep{tang2024belted} & \ding{55} & \ding{51} & \ding{55} & Potential \\
        GenSDR (ours) & \ding{51} & \ding{51} & \ding{51} & \ding{51} \\
        \bottomrule
    \end{tabular}
    \label{tab: method_comparison}
\end{table}

Our main contributions are threefold. First, we formulate nonlinear SDR through the nested structure of the velocity field induced by stochastic interpolation, and, based on this insight, develop the GenSDR method. We establish a rigorous and general population-level theory demonstrating the exhaustiveness of our estimators. Second, we analyze the smoothness properties of the conditional velocity field under the low-dimensional sufficient structure, which may be of independent interest, and prove sample-level convergence of the estimated conditional distribution to the true conditional distribution. To the best of our knowledge, this is the first work to provide distributional consistency guarantees for empirical nonlinear SDR estimators. Third, we extend GenSDR to handle non-Euclidean responses, thereby broadening its applicability. A comparative overview is provided in Table \ref{tab: method_comparison}, which clearly situates our method within the broader methodological landscape.

The remainder of this paper is organized as follows. In Section \ref{sec: preliminaries}, we briefly review neural networks and conditional stochastic interpolation models. Section \ref{sec: methodology} introduces GenSDR for Euclidean responses and establishes its population-level effectiveness. 
In Section \ref{sec: illustrative_example}, we use a concrete example to illustrate the theoretical intuition and logical reasoning behind our method, followed by an error analysis. In Section \ref{sec: empirical_consistency}, we establish the convergence rate of the GenSDR estimator in general settings. Section \ref{sec: beyond_euclidean} extends our method to non-Euclidean responses. Numerical experiments in Section \ref{sec: numerical_experiments} demonstrate the competitiveness of GenSDR relative to existing methods. Section \ref{sec: conclusion} concludes the paper. The detailed proofs for technical results are deferred to the Supplementary Material.

\section{Preliminaries}\label{sec: preliminaries}

\subsection{Neural networks}

Neural networks constitute a pivotal class of models in modern machine learning and are widely used for approximating complex functions \citep{shen2020deep, siegel2023optimal, havrilla2024understanding}. In this paper, we focus on a fundamental architecture, the multi-layer perceptron (MLP), which is structured as a stack of linear transformations and nonlinear activations. Mathematically, a function $f$ implemented by an MLP of depth $L$ can be written as
$$
f(x)
= \phi_{L+1} \circ \sigma_L \circ \phi_L \circ \cdots \circ \sigma_1 \circ \phi_1(x),
$$
where $\phi_i(x) = W_i x + b_i$, with $W_i$ a matrix of size $d_i \times d_{i-1}$ and $b_i \in \mathbb{R}^{d_i}$, and all activation functions are taken to be the rectified linear unit (ReLU), i.e., $\sigma_i(x) = \max(x, 0)$ applied elementwise, for $i = 1,\dots,L+1$. The width of the network is defined as $\max_{1 \le i \le L} d_i$.

An appealing property of MLPs is that a truncated neural network function can be represented by a deeper neural network, which is useful for enforcing boundedness, as it is typically required that estimators not to grow unboundedly. Throughout, we assume that the range of any neural network function is uniformly bounded above and below. Consequently, a neural network function class is characterized by its input and output dimensions, depth, width, total number of parameters, and uniform upper and lower bounds on its outputs.

\subsection{Local Sobolev space}

To analyze the empirical behavior of a nonparametric estimator, it is crucial to quantify how well the unknown target function can be approximated by a given function class, such as a neural network class \citep{gyorfi2002distribution, dytso2022conditional, shen2022approximation}. The approximation error depends on the smoothness of the true function. As we will show in subsequent sections, the conditional velocity field often satisfies a Lipschitz-type regularity condition. This motivates our focus on Sobolev-type function spaces. The Sobolev space in Definition~\ref{def: sobolev_space} comprises functions with bounded derivatives up to a prescribed order, while the local Sobolev space in Definition~\ref{def: local_sobolev_space} additionally controls the growth of these derivatives as the domain expands.

\begin{definition}[Sobolev space]\label{def: sobolev_space}
Let $\beta\in\mathbb{N}$ and $\Omega\subseteq\mathbb{R}^d$. The Sobolev space $\mathcal{W}^{\beta, \infty}(\Omega)$ is defined by
$$
\mathcal{W}^{\beta, \infty}(\Omega)=\left\{f: \Omega\to\mathbb{R}, \|f\|_{\infty}<\infty, \|D^{\alpha}f\|_{\infty}<\infty \text{ for all } \alpha\in\mathbb{N}^d \text{ with } \|\alpha\|_1\le \beta\right\}.
$$
Furthermore, for any $f\in\mathcal{W}^{\beta, \infty}(\Omega)$, we define the Sobolev norm $\|\cdot\|_{\mathcal{W}^{\beta, \infty}(\Omega)}$ as
$$
\|f\|_{\mathcal{W}^{\beta, \infty}(\Omega)}=\max_{0\le \|\alpha\|_1\le \beta}\|D^{\alpha}f\|_{\infty}.
$$
\end{definition}

\begin{definition}[Local Sobolev space]\label{def: local_sobolev_space}
Let $\beta\in\mathbb{N}$. The Local Sobolev space $\mathcal{W}_{\mathrm{Loc}}^{\beta, \infty}(\mathbb{R}^d, B)$ is defined by
$$
\begin{aligned}
\mathcal{W}_{\mathrm{Loc}}^{\beta, \infty}(\mathbb{R}^d, B)=\big\{f: \mathbb{R}^d\to\mathbb{R},\ & h(x)=f_{|(-u, u)^d}(2ux-u1_d)\in \mathcal{W}^{\beta, \infty}((0, 1)^d), \\
&  \|h\|_{\mathcal{W}^{\beta, \infty}((0, 1)^d)}\le B(u+1), \text{for any } u>0\big\}.
\end{aligned}
$$
Here, $1_d$ denotes the $d$-dimensional vector with entries all being 1, and $f_{|\Omega}$ means the restriction of function $f$ to the domain $\Omega$.
\end{definition}

\subsection{Conditional stochastic interpolation}

A conditional stochastic interpolation model \citep{albergo2023building, huang2023conditional} is a flow-based generative model that learns a conditional distribution by constructing a suitable interpolation path. Specifically, let $\mathcal{I}(y_0, y_1, t)$ be a continuous interpolant connecting an initial point $y_0$ and a terminal point $y_1$ as $t$ ranges from $0$ to $1$, such that $\mathcal{I}(y_0, y_1, 0) = y_0$ and $\mathcal{I}(y_0, y_1, 1) = y_1$. Given $X$, we construct the interpolation
$$
Y_t = \mathcal{I}(\eta, Y, t),
$$
where $\eta \sim \mathcal{N}(0, I_{d_y})$ is a $d_y$-dimensional standard Gaussian random vector, independent of $(X, Y)$. Define the time-dependent conditional velocity field
\begin{equation}\label{eqn: raw_conditional_velocity_field}
b_0(x, y, t)
= \mathbb{E}\bigl[\partial_t \mathcal{I}(\eta, Y, t) \,\big|\, Y_t = y, X = x\bigr].
\end{equation}
Let $p(x, \cdot, t)$ denote the probability density function of $Y_t | X = x$ for $t \in [0, 1]$. A key property is that, for any fixed $x$, the family of densities $\{p(x, \cdot, t) : t \in [0, 1]\}$ solves the following transport equation with respect to $\{\rho(\cdot, t): \mathbb{R}^{d_y} \to \mathbb{R},\, t \in [0, 1]\}$ \citep{liu2022flow, lipman2023flow},
\begin{equation}\label{eqn: trainsport_equation}
\partial_t \rho(y, t) + \nabla_y \cdot \bigl[b_0(x, y, t)\,\rho(y, t)\bigr] = 0.
\end{equation}
Here, $\nabla_y \cdot$ denotes the divergence operator with respect to $y$. The transport equation \eqref{eqn: trainsport_equation} is closely associated with the ordinary differential equation (ODE)
\begin{equation}\label{eqn: true_ode}
\mathrm{d}Z_t = b_0(x, Z_t, t)\,\mathrm{d}t,
\end{equation}
for fixed $x$. Consequently, if Eq.~\eqref{eqn: trainsport_equation} admits a unique solution, then the terminal distribution of the ODE \eqref{eqn: true_ode} at $t = 1$, with initial condition $Z_0 \sim \mathcal{N}(0, I_{d_y})$, coincides with the conditional distribution of $Y | X = x$ \citep{gao2024gaussian}. The uniqueness of the solution to Eq.~\eqref{eqn: trainsport_equation} (equivalently, Eq.~\eqref{eqn: true_ode}) plays a crucial role in our theoretical development at both the population and sample levels; see Supplementary Material Section A.1 for further details.

\subsection{Notations}

For a vector $v \in \mathbb{R}^d$, we use $\|v\|_1$, $\|v\|_2$, and $\|v\|_{\infty}$ to denote its $\ell_1$-norm, Euclidean norm, and $\ell_{\infty}$-norm, respectively; $v_{(j)}$ denotes its $j$-th component for $j = 1, \dots, d$. For a vector-valued function $f: \mathbb{R}^{d_1} \to \mathbb{R}^{d_2}$, we write $f_{(j)}$ for its $j$-th coordinate function, $j = 1, \dots, d_2$. For a function $f(x, t)$ with $t \in \mathbb{R}$, $\partial_t f(x, t)$ denotes the partial derivative of $f$ with respect to $t$. For a scalar function $\alpha_t$ of $t \in \mathbb{R}$, we write $\dot{\alpha}_t$ for its derivative with respect to $t$.

We use $\mathds{1}(\cdot)$ for the indicator function, $1_d$ for the $d$-dimensional vector of ones, and $I_d$ for the $d$-dimensional identity matrix. For a random vector $X$, $\mathrm{supp}(X)$ denotes the support of $X$. We use generic symbols $\mathcal{F}_1$ and $\mathcal{F}_2$ to denote function classes over which optimization is performed; these will be specified in the statements of theorems. We use generic symbols $g^*$ and $R^*$ (and their empirical counterparts $\hat{g}$ and $\hat{R}$) to denote minimizers arising from our method, with their precise definitions depending on the context, as detailed in Sections~\ref{sec: methodology}, \ref{sec: illustrative_example}, \ref{sec: empirical_consistency}, and \ref{sec: beyond_euclidean}.

\section{Methodology}\label{sec: methodology}

\subsection{GenSDR}

We aim to recover an equivalent sufficient transformation to the target $R_0$ that induces the sufficient representation. Eq.~\eqref{eqn: nonlinear_sdr_representation} indicates that the two conditional distributions $Y|X$ and $Y|R_0(X)$ are the same. To this end, we leverage recent advances in generative models with conditional distribution as the learning target \citep{zhou2023deep, yang2025conditional, zhao2025conditional}, in particular the stochastic interpolation framework. By introducing a temporal dimension, the conditional distribution is represented via a conditional velocity field, which is substantially easier to handle. Crucially, this conditional velocity field preserves the low-dimensional structure, which enables the estimation of $R_0$. Throughout, we assume that $Y| X=x$ admits a probability density function for any $x\in\mathrm{supp}(X)$, where $\mathrm{supp}(X)$ denotes the support of $X$. This assumption is imposed for theoretical convenience; certain advanced flow-matching models can accommodate discrete responses \citep{holderrieth2025generator}.

We first illustrate how the low-dimensional structure in formula \eqref{eqn: nonlinear_sdr_representation} is encoded in the conditional velocity field. Let $\mathcal{I}(y_0, y_1, t)$ be an interpolant and denote by $b_0(x, y, t)$ the conditional velocity field induced by $\mathcal{I}$ (see Eq.~\eqref{eqn: raw_conditional_velocity_field}). By construction, $Y_t$ is generated from $\eta$ and $Y$. Since $\eta$ is independent of $(X, Y)$, it follows that $(Y, Y_t)\perp\!\!\!\perp X| R_0(X)$. Consequently, the velocity field $b_0(x, y, t)$ depends on $x$ only through $R_0(x)$. Lemma~\ref{lem: low_dim_veclocity} formalizes this property.

\begin{lemma}\label{lem: low_dim_veclocity}
Suppose that for any $t\in [0, 1]$, $\mathcal{I}(y_0, y_1, t)$ is injective in either $y_0$ or $y_1$. Then, we have
$$
b_0(x, y, t)\equiv \mathbb{E}\left[\partial_t\mathcal{I}(\eta, Y, t)|Y_t=y, R_0(X)=R_0(x)\right].
$$
\end{lemma}

The injectivity condition in Lemma~\ref{lem: low_dim_veclocity} is mild; for example, the straight-line interpolant $\mathcal{I}_{\mathrm{sl}}(y_0, y_1, t)=(1-t)y_0+ty_1$ satisfies this condition and is widely used in generative modeling (see, for instance, \cite{liu2022flow, kornilov2024optimal}). Lemma~\ref{lem: low_dim_veclocity} implies the existence of a time-dependent function $g_0$ such that
\begin{equation}\label{eqn: conditional_velocity_identity}
b_0(x, y, t)\equiv g_0(R_0(x), y, t).
\end{equation}
Identity \eqref{eqn: conditional_velocity_identity} suggests that $R_0$ can be learned jointly with $g_0$. Specifically, if $g_0$ is sufficiently \emph{regular}, any incorrect transformation $R$ cannot reproduce the true conditional velocity field via $g_0(R(X), y, t)$. Motivated by this, we propose to estimate $R_0$ and $g_0$ by minimizing the following squared loss,
\begin{equation}\label{eqn: population_loss}
g^*, R^*=\mathop{\mathrm{argmin}}_{g\in\mathcal{F}_1,\, R\in\mathcal{F}_2}\int_0^1\mathbb{E}\bigl\|\partial_t\mathcal{I}(\eta, Y, t)-g(R(X), Y_t, t)\bigr\|_2^2\,\mathrm{d}t,
\end{equation}
where $\mathcal{F}_1$ and $\mathcal{F}_2$ are function classes that ensure, at a minimum, the integrability of $\mathbb{E}\|g(R(X), Y_t, t)\|_2^2$ over $t\in[0,1]$. The estimator $R^*$ serves as an estimate of $R_0$ and is expected to encode all information contained in $\mathcal{G}_{Y| X}$. We refer to this method as GenSDR, emphasizing its utilization of generative models. The minimizer of \eqref{eqn: population_loss} is generally not unique; indeed, any bijective transformation of $R_0$ yields an equivalent $g_0$. Consequently, the pair $(g^*, R^*)$ should be regarded as an element of the set of minimizers of \eqref{eqn: population_loss}.

\subsection{Population-level validity}

We now address the key question: Is the learned transformation $R^*$ sufficient? The notion of \emph{sufficiency} here is nuanced. In the strictest sense, sufficiency requires
$$
\sigma(R^*(X))=\sigma(R_0(X))=\mathcal{G}_{Y| X},
$$
corresponding to the ideal case in which no relevant information is lost or redundantly included. As discussed in \cite{lee2013general}, the equality $\sigma(R^*(X))=\mathcal{G}_{Y| X}$ implies that $R^*$ is both unbiased and exhaustive. While many existing methods focus on unbiasedness $\sigma(R^*(X)) \subseteq \mathcal{G}_{Y| X}$ \citep{lee2013general, chen2024deep}, GenSDR is primarily designed to ensure \emph{exhaustiveness}; that is,
$$
\mathcal{G}_{Y| X}\subseteq\sigma(R^*(X)).
$$

Non-exhaustiveness is particularly problematic. Consider two candidate transformations $R_1$ and $R_2$ mapping $X$ to latent spaces of the same dimension, with
$$
\sigma(R_1(X))\subseteq\mathcal{G}_{Y| X}\subseteq\sigma(R_2(X)).
$$
In this case, $R_2$ is preferable, as $R_2(X)$ preserves all relevant information without increasing dimensionality, whereas using $R_1$ for downstream tasks may lead to substantial errors due to information loss. For instance, consider $Y=X_{(1)}\mathds{1}(X_{(1)}>0)+\varepsilon$ where $X_{(1)}$ denotes the first component of $X$ and is non-degenerate on $\mathbb{R}$, with $\varepsilon\perp\!\!\!\perp X$. Let $R_1(X)=X_{(1)}\mathds{1}(X_{(1)}>1)$ and $R_2(X)=X_{(1)}$. Noticeably, using $R_1$ as a predictor for $Y$ discards substantial information whenever $X_{(1)}\in (0, 1]$, yielding a significant and unavoidable prediction error. In contrast, $R_2$ retains the full signal contained in $X_{(1)}\mathds{1}(X_{(1)}>0)$; although it may introduce slight redundancy, it preserves all relevant information without increasing the underlying latent dimensionality. On the other hand, the flexibility of $g_0$ may prevent $R^*$ from being the most parsimonious representation; it is plausible that $Y\perp\!\!\!\perp X| R^*(X)$ holds, yet $R^*$ is not related to $R_0$ by a bijection. Nevertheless, exhaustiveness remains a highly nontrivial property that existing methods typically struggle to guarantee.

Define
$$
\nu_0=\int_0^1\mathbb{E}\bigl\|\partial_t\mathcal{I}(\eta, Y, t)-b_0(X, Y_t, t)\bigr\|_2^2\,\mathrm{d}t.
$$
Theorem~\ref{thm: pl_effectiveness} shows that, under suitable regularity conditions, the $\sigma$-field generated by the population level minimizer $R^*$ fully contains the target $\mathcal{G}_{Y| X}$, which establishes exhaustiveness.

\begin{theorem}\label{thm: pl_effectiveness}
Assume that
\begin{enumerate}[label=(\roman*)]
    \item for any $t\in [0, 1]$, $\mathcal{I}(y_0, y_1, t)$ is injective in either $y_0$ or $y_1$;
    \item $g_0(\cdot, \cdot, \cdot)$ is continuous over $\mathbb{R}^d\times \mathbb{R}^{d_y}\times [0, 1]$, $R_0(\cdot)$ is continuous over $\mathrm{supp}(X)$, $\mathcal{F}_1\subseteq\mathcal{C}(\mathbb{R}^{q}\times\mathbb{R}^{d_y}\times [0, 1], \mathbb{R}^{d_y})$ and $\mathcal{F}_2\subseteq\mathcal{C}(\mathbb{R}^{d_x}, \mathbb{R}^q)$ with some integer $q\ge d$, such that there exists $(g^{\dagger}, R^{\dagger})\in\mathcal{F}_1\times \mathcal{F}_2$ satisfying $g_0(R_0(X), Y_t, t)=g^{\dagger}(R^{\dagger}(X), Y_t, t)$ a.s. $(\eta, X, Y)$ and a.e. $t\in [0, 1]$;
    \item $\nu_0$ is finite;
    \item for any $t\in [0, 1]$, $(x, y)\in \mathrm{supp}(X, Y_t)$ implies the existence of a positive scaler $\epsilon_t$ such that the Lebesgue measure of the joint set $\{s: (x, y)\not\in\mathrm{supp}(X, Y_s)\}\cap\{s: |s-t|\le \epsilon_t\}$ is 0;
    \item for any $t\in [0, 1]$ and $x\in\mathrm{supp}(X)$, $(x, y)\in\mathrm{supp}(X, Y_t)$ implies $y\in\mathrm{supp}(Y_t|X=x)$;
    \item for any $x\in\mathrm{supp}(X)$, the ODE induced by $b_0(x, y, t)$ with Gaussian initialization admits a unique solution, such that the solution at any time $t\in[0, 1]$ follows the distribution of $Y_t|X=x$.
\end{enumerate}
Then, for each pair $(g^*, R^*)$ defined in Eq. \eqref{eqn: population_loss}, we have $\mathcal{G}_{Y|X}\subseteq\sigma(R^*(X))$.
\end{theorem}

The conditions in Theorem~\ref{thm: pl_effectiveness} are fairly general. Condition \textit{(i)} is inherited directly from Lemma~\ref{lem: low_dim_veclocity}. Condition \textit{(ii)} does not require $R^*(X)$ to be exactly $d$-dimensional, thus permitting model misspecification. The assumption that $g(\cdot, \cdot, t)$ is continuous in $t\in[0,1]$ may be somewhat restrictive, as it disregards potential temporal singularities of the conditional velocity field \citep{jiao2024convergence, pmlr-v267-zhou25l}. Condition \textit{(iii)} ensures that the minimization problem is non-degenerate and admits a nontrivial minimizer. When the interpolant is chosen to be linear (see Section~\ref{sec: illustrative_example}), this can be guaranteed if $\|Y\|_2$ has a finite second moment. Conditions \textit{(iv)} and \textit{(v)} are technical conditions which regulate the density of the distribution of $(X, Y_t)$. Together with condition \textit{(ii)}, the two conditions ensure that equality almost everywhere implies equality everywhere. Condition \textit{(vi)} is essential for relating the conditional velocity field to the conditional distribution, and relies on the smoothness of the conditional velocity field \citep{bris2008existence}. Although these conditions appear abstract, we provide a simple yet illustrative example in Section~\ref{sec: illustrative_example} demonstrating that they can be satisfied in a nontrivial and concrete setting.

Theorem~\ref{thm: pl_effectiveness} thus establishes the exhaustiveness of GenSDR under appropriate conditions, highlighting its ability to comprehensively capture information in $\mathcal{G}_{Y| X}$. In contrast, 
prominent RKHS-based nonlinear SDR methods, such as GSIR and GSAVE, are guaranteed to be unbiased but not necessarily exhaustive; see Theorems 7 and 8 in \cite{lee2013general}. Regarding deep learning-based nonlinear SDR methods, while GMDDNet is proved to achieve unbiasedness only, the capability of DDR to fully recover $\mathcal{G}_{Y| X}$ remains unclear.
Moreover, both our method and BENN \citep{tang2024belted} can be viewed as addressing nonlinear SDR in an augmented space that incorporates multiple perspectives of the response $Y$ simultaneously. We argue that the robustness of the learned transformation $R^*$ to different augmentations of $Y$ is central to its exhaustiveness.

\section{An illustrative example}\label{sec: illustrative_example}

This section presents the rationale of GenSDR through a simple Gaussian example. We derive the explicit conditional velocity field, which reveals that a sufficient transformation is learnable within our framework. We also describe the corresponding sample estimator and establish its non-asymptotic convergence rate in terms of estimation error. This warm-up example provides an insightful understanding of the mechanism inside GenSDR.

\subsection{Closed form of the conditional velocity field}

Consider the model
$$
Y = f_0(X)\cdot\varepsilon,
$$
where $\varepsilon$ follows a standard $d_y$-dimensional Gaussian distribution and is independent of the covariate $X$, and $f_0: \mathbb{R}^{d_x}\to\mathbb{R}$ is an unknown nondegenerate function (that is, $f_0(x)$ is not a constant). We distinguish $f_0$ from $R_0$ since $|f_0(X)|$ generates the central $\sigma$-field, due to the symmetry of $\varepsilon$. It is important to note that, in this setting, the object of estimation departs from the classical paradigm of the conditional mean and instead centers on the conditional covariance. For the interpolation path, we adopt the linear interpolant
$$
\mathcal{I}_{\mathrm{lin}}(y_0, y_1, t) = \alpha_t y_0 + \beta_t y_1,
$$
where $\alpha_t$ and $\beta_t$ are continuously differentiable in $t$ and satisfy the boundary conditions $\alpha_0 = \beta_1 = 1$ and $\alpha_1 = \beta_0 = 0$. Consequently, the conditional velocity field given $X = x$ is
$$
b_0(x, y, t) = \mathbb{E}\bigl(\dot{\alpha}_t \eta + \dot{\beta}_t Y \,\big|\, Y_t = y, X = x\bigr),
$$
where $Y_t = \alpha_t \eta + \beta_t Y$ given $X = x$. Under this linear interpolation, the conditional velocity field $b_0(x,y,t)$ admits a closed form as stated in Proposition~\ref{prop: warm_up_cvf}. This proposition clarifies that sufficient representation $|f_0(X)|$ fundamentally characterizes the dependency of the conditional velocity field on $X$.

\begin{proposition}\label{prop: warm_up_cvf}
Assume that $\alpha_t^2+\beta_t^2\ne 0$ for any $t\in [0, 1]$, and $0\not\in \{f_0(x): x\in \mathbb{R}^{d_x}\}$. Then, for any $t\in [0, 1]$, we have
\begin{equation}\label{eqn: warm_up_cvf}
b_0(x, y, t)=\frac{\dot{\alpha}_t\alpha_t+\dot{\beta}_t\beta_tf_0(x)^2}{\alpha_t^2+\beta_t^2f_0(x)^2}y.
\end{equation}
\end{proposition}

The conditions in Proposition~\ref{prop: warm_up_cvf} are consistent with the injectivity condition in Lemma~\ref{lem: low_dim_veclocity}. Moreover, the requirement that $f_0(x)\neq 0$ is natural; otherwise, $Y$ would be degenerate and fail to admit a probability density function. Together with the condition $\alpha_t^2 + \beta_t^2 \neq 0$, this implies $\alpha_t^2 + \beta_t^2 f_0(x)^2 > 0$, ensuring that the right-hand side of Eq.~\eqref{eqn: warm_up_cvf} is well-defined. Under Gaussian setup, the sufficient transformation $R_0$ (which can be chosen as $|f_0|$) is tractable, illustrating an advantage of using the conditional velocity field.

Given Eq.~\eqref{eqn: warm_up_cvf}, we can rewrite $b_0(x, y, t)$ as $g_0(f_0(x), y, t)$, where
$$
g_0(z, y, t)
= \frac{\dot{\alpha}_t \alpha_t + \dot{\beta}_t \beta_t z^2}{\alpha_t^2 + \beta_t^2 z^2}\, y.
$$
Proposition~\ref{prop: warm_up_continuity} below specifies the smoothness properties of the proxy function $g_0$, which are crucial for analyzing the empirical estimator defined subsequently. 

\begin{proposition}\label{prop: warm_up_continuity}
Assume that $\alpha_t$ and $\beta_t$ are twice continuously differentiable, and $\alpha_t^2+\beta_t^2\ne 0$ for any $t\in [0, 1]$. Then, for every $j\in \{1, \dots, d_y\}$, any $A>0$, and any constants $c_1, c_2$ satisfying $c_2> c_1> 0$, we have
\begin{enumerate}[label=(\roman*)]
    \item $\sup_{t\in [0, 1], |z|\in [c_1, c_2]}\sup_{\|y\|_{\infty}\le A}|g_{0, (j)}(z, y, t)|\le c^*A$;
    \item $\sup_{t\in [0, 1], |z|\in [c_1, c_2]}\sup_{\|y\|_{\infty}\le A}|\partial_tg_{0, (j)}(z, y, t)|\le c^*A$;
    \item $\sup_{t\in [0, 1], |z|\in [c_1, c_2]}\sup_{\|y\|_{\infty}\le A}|\partial_zg_{0, (j)}(z, y, t)|\le c^*A$;
    \item $\sup_{t\in [0, 1], |z|\in [c_1, c_2]}\sup_{y\in\mathbb{R}^{d_y}}\|\nabla_yg_{0, (j)}(z, y, t)\|_{\infty}\le c^*$.
\end{enumerate}
Here, $c^*$ is a constants which does not depend on $d_y$.
\end{proposition}

Recall that our goal is to learn a sufficient transformation equivalent to $|f_0|$, which captures all information in $X$ relevant for $Y$. At the population level, the loss function derived from Eq.~\eqref{eqn: population_loss} is
\begin{equation}\label{eqn: warm_up_population_loss}
g^*, f^*
= \mathop{\mathrm{argmin}}_{g\in \mathcal{F}_1,\, f\in\mathcal{F}_2}
\int_0^1
\mathbb{E}\bigl\|
\dot{\alpha}_t \eta + \dot{\beta}_t Y - g\bigl(f(X), Y_t, t\bigr)
\bigr\|_2^2\,\mathrm{d}t.
\end{equation}
In this warm-up setting, the population-level validity can be verified explicitly, as stated in Corollary~\ref{coro: warm_up_population_validity}. 
Note that the velocity field here does not exhibit a singularity in $t$ as $t\to 1$, provided $|f_0(x)|$ is bounded away from zero. In addition, the uniqueness of the ODE solution is ensured by the Lipschitz continuity of the conditional velocity field \citep{teschl2012ordinary}, see Supplementary Material Section A.1 for a detailed discussion.

\begin{corollary}\label{coro: warm_up_population_validity}
Assume that
\begin{enumerate}[label=(\roman*)]
    \item there exists an absolute constant $\epsilon>0$ such that $\alpha_t, \beta_t$ are continuous differentiable and $\alpha_t^2+\beta_t^2\ne 0$ for any $t\in (-\epsilon, 1+\epsilon)$;
    \item $(g_0, f_0)\in \mathcal{F}_1\times \mathcal{F}_2$, where $\mathcal{F}_1\subseteq\mathcal{C}(\mathbb{R}\times\mathbb{R}^{d_y}\times [0, 1], \mathbb{R}^{d_y})$ and $\mathcal{F}_2\subseteq\mathcal{C}(\mathbb{R}^{d_x}, \mathbb{R})$;
    \item there exist positive absolute constants $C_1$ and $C_2$ such that $\{|f_0(x)|: x\in \mathbb{R}^{d_x}\}\subseteq [C_1, C_2]$.
\end{enumerate}
Then, for each pair $(g^*, f^*)$ defined in Eq. \eqref{eqn: warm_up_population_loss}, we have $\mathcal{G}_0=\sigma(|f_0(X)|)\subseteq\sigma(f^*(X))$.
\end{corollary}

\subsection{Non-asymptotic empirical convergence rate}

At the sample level, we construct estimators for $g_0$ and $|f_0|$ based on observations $\{(X_1, Y_1), \dots, (X_n, Y_n)\}$. To realize the stochastic interpolation, we generate auxiliary samples $\{\eta_1, \dots, \eta_n\}$ and $\{t_1, \dots, t_n\}$ from a standard Gaussian distribution on $\mathbb{R}^{d_y}$ and the uniform distribution on $[0, 1]$, respectively. The empirical estimators $\hat{g}$ and $\hat{f}$ are defined as
$$
\hat{g}, \hat{f}
= \mathop{\mathrm{argmin}}_{g\in \mathcal{F}_1,\, f\in\mathcal{F}_2}
\frac{1}{n}\sum_{i=1}^n
\bigl\|
\dot{\alpha}_{t_i}\eta_i + \dot{\beta}_{t_i}Y_i
- g\bigl(f(X_i), Y_{i, t_i}, t_i\bigr)
\bigr\|_2^2,
$$
where $\mathcal{F}_1$ and $\mathcal{F}_2$ are function classes (implemented by neural networks in practice), and $Y_{i, t_i} = \alpha_{t_i}\eta_i + \beta_{t_i}Y_i$ for $i=1, \dots, n$.

Given this estimation paradigm, a key aspect is to evaluate the performance of the learned representation $\hat{f}$. In linear SDR, one typically measures the distance between the true and estimated dimension reduction matrices \citep{xia2002adaptive, xu2025neural}. In nonlinear scenarios, however, the true representational function (that is, the sufficient transformation) is non-identifiable and its relationship to the identifiable central $\sigma$-field is more intricate. Particularly, Theorem~\ref{thm: pl_effectiveness} and Corollary~\ref{coro: warm_up_population_validity} indicate that $\sigma(f^*(X))$ can be slightly larger than $\mathcal{G}_{Y| X}$, making a direct comparison between $\hat{f}$ and $|f_0|$ under a common distance metric difficult to interpret. We therefore focus on recovering the conditional distribution rather than the representational function itself, which is more relevant and appropriate, and simultaneously suggests the generation efficiency under low-dimensional structures. This evaluation is conducted via the standard sampling phase \citep{albergo2023building}.

Given $\hat{g}$ and $\hat{f}$, we consider the ODE
$$
\mathrm{d}\tilde{Y}_t
= \hat{g}\bigl(\hat{f}(x), \tilde{Y}_t, t\bigr)\,\mathrm{d}t,
\quad \tilde{Y}_0 \sim \eta,
$$
for $t\in[0, 1]$ and any fixed $X = x$. Incorporating time discretization (Euler method), a more realistic sampling scheme is
$$
\mathrm{d}\hat{Y}_t
= \hat{g}\bigl(\hat{f}(x), \hat{Y}_{\bar{t}_k}, \bar{t}_k\bigr)\,\mathrm{d}t,
\quad t\in[\bar{t}_k, \bar{t}_{k+1}],\ k=0, \dots, K-1,
\quad \hat{Y}_0 \sim \eta,
$$
for $t\in[0, 1]$ and $X = x$, where $\{\bar{t}_0, \dots, \bar{t}_K\}$ is a prescribed time grid with $\bar{t}_0 = 0$ and $\bar{t}_K = 1$. This dual-level ODE formulation follows \cite{gao2024convergence} and \cite{jiao2024convergence}. Denote by $\tilde{\pi}_t(x)$ and $\hat{\pi}_t(x)$ the distributions of $\tilde{Y}_t| \hat{f}(X)=\hat{f}(x)$ and $\hat{Y}_t| \hat{f}(X)=\hat{f}(x)$, respectively.

We use the 2-Wasserstein distance \citep{kantorovich1960mathematical, shen2018wasserstein} to quantify the discrepancy between probability distributions. For distributions $\lambda_1$ and $\lambda_2$ with finite second moments, the 2-Wasserstein distance $W_2(\lambda_1, \lambda_2)$ is defined as
$$
W_2(\lambda_1, \lambda_2)
= \Biggl(
\inf_{\lambda\in\Pi(\lambda_1, \lambda_2)}
\int \|z_1 - z_2\|_2^2\,\mathrm{d}\lambda(z_1, z_2)
\Biggr)^{1/2},
$$
where $\Pi(\lambda_1, \lambda_2)$ denotes the set of all couplings of $\lambda_1$ and $\lambda_2$. In nonlinear SDR, we thus study the estimation error
$$
\mathbb{E}\bigl[W_2(\pi_t(X), \hat{\pi}_t(X))\bigr]
= \int \mathbb{E}\bigl[W_2(\pi_t(x), \hat{\pi}_t(x))\bigr]\,\mathrm{d}\mathbb{P}_X(x),
$$
where $\pi_t(x)$ denotes the conditional distribution of $Y_t=\mathcal{I}_{\mathrm{lin}}(\eta, Y, t)$, the intermediate state of the stochastic interpolation given $X = x$, and $\mathbb{P}_X$ is the probability measure of the covariates $X$. Another closely related ODE is
\begin{equation}\label{eqn: warm_up_true_ode}
\mathrm{d}\bar{Y}_t
= g_0\bigl(f_0(x), \bar{Y}_t, t\bigr)\,\mathrm{d}t,
\quad t\in[0,1],
\quad \bar{Y}_0 \sim \eta,
\end{equation}
since we will reach the estimation error between conditional distributions through that between conditional velocity fields. Under standard regularity conditions (for instance, the ODE in Eq.~\eqref{eqn: warm_up_true_ode} has a unique solution trajectory), $\bar{Y}_t$ and $Y_t$ share the same distribution for any $t\in[0, 1]$.

To make it explicit, for the final timestep $t=1$, $\pi_1(X)$ represents the target conditional distribution $Y|X$. Correspondingly, $\hat\pi_1(X)$ is the learned conditional distribution $\hat{Y_1}| \hat{f}(X)$, generated based on the estimated low dimensional sufficient representation $\hat f(X)$. Theorem~\ref{thm: warm_up_estimation_error} shows that the estimation error can be controlled under certain assumptions, with an explicit convergence rate present.

\begin{theorem}\label{thm: warm_up_estimation_error}
Assume that
\begin{enumerate}[label=(\roman*)]
    \item $\alpha_t$ and $\beta_t$ are twice continuously differentiable, and $\alpha_t^2+\beta_t^2\ne 0$ for all $t\in [0, 1]$;
    \item there exist positive absolute constants $C_1$ and $C_2$ such that $\{f_0(x): x\in\mathbb{R}^{d_x}\}\subseteq [C_1, C_2]$;
    \item $f_0\in\mathcal{W}_{\mathrm{Loc}}^{1, \infty}(\mathbb{R}^{d_x}, C_2)$ and $\|X\|_{\infty}$ follows a sub-Gaussian distribution.
\end{enumerate}
Let $\mathcal{F}_1=\mathcal{F}_{1, \mathrm{NN}}\cap \mathcal{F}_{\mathrm{Lip}, \Lambda_n}$. Here, $\mathcal{F}_{1, \mathrm{NN}}$ a neural network function class with an input dimensionality $d_y+2$, output dimensionality $d_y$, depth $L_1$, width $M_1$, uniform upper bound $\bar{\delta}_{1, n}=(\log n)^{(1+\kappa)/2}$, uniform lower bound $\underline{\delta}_{1, n}=-(\log n)^{(1+\kappa)/2}$, where $\kappa\in (0, 1)$ is an arbitrarily fixed real number, and $\Lambda_n=(\log n)^{(1+\kappa)/2}$. Let $\mathcal{F}_2=\mathcal{F}_{2, \mathrm{NN}}$ be a neural network function class with an input dimensionality $d_x$, output dimensionality 1, depth $L_2$, width $M_2$, uniform upper bound $\bar{\delta}_{2, n}=\log\log n$, uniform lower bound $\underline{\delta}_{2, n}=0$. Let $\Delta=\max(\bar{t}_1-\bar{t}_0, \dots, \bar{t}_K-\bar{t}_{K-1})\le n^{-1}$. Then, for any $n\ge 2$, we have
$$
\mathbb{E}\left[W_2(\pi_1(X), \hat{\pi}_1(X))\right]\le c^*n^{-\frac{1}{d^*+2}}(\log n)^{2+\kappa}\exp\{(\log n)^{(1+\kappa)/2}\},
$$
where $d^*=\max(d_x, d_y+2)$, and $c^*$ is a constant which does not depend on $n$.
\end{theorem}

In Theorem~\ref{thm: warm_up_estimation_error}, we assume $f_0$ is positive and therefore do not distinguish between $f_0$ and $|f_0|$; this mild modification simplifies the analysis. Theorem~\ref{thm: warm_up_estimation_error} characterizes the empirical performance of $\hat{f}$ from the perspective of conditional distribution recovery via a generative model, yielding a more fundamental result for sufficient representation learning than Theorem~3 in \cite{tang2024belted}, which only establishes convergence of a method-specific excess risk (analogous to the estimation error of the conditional velocity field in our framework). Moreover, compared to the standard convergence rate $n^{-1/(d_x+d_y+2)}$ for conditional distribution estimation \citep{li2022minimax, ICLR2025_a5e2853b}, the derived error rate $n^{-1/(d^*+2)}$ exhibits potential fast convergence. The enhanced convergence rate is a direct consequence of the advantages offered by sufficient dimension reduction on predictor $X$ from $d_x$ to $1$, the effect of which is most pronounced in high-dimensional $Y$ settings. Essentially, Theorem~\ref{thm: warm_up_estimation_error}  characterizes the convergence property of GenSDR under the condition that the conditional velocity field possesses ideal properties such as smoothness and non-singularity. The next section will focus on the analysis of the convergence properties of GenSDR in more general cases.


\section{Error analysis for general cases}\label{sec: empirical_consistency}

In this section, we define empirical estimators for $g_0$ and $R_0$ in generic settings and study the consistency of the estimator of $R_0$. It is crucial to understand the smoothness of the conditional velocity field more thoroughly and to control its potential blow-up as $t\to 1$. Such behavior is frequently observed in diffusion and flow-matching models \citep{yang2024lipschitz, zhang2024tackling}; under the linear interpolant, the velocity field is tightly linked to the score function via Tweedie's formula. We primarily show that our estimator of $R_0$ is consistent at the level of conditional distributions, while obtaining a fast convergence rate requires more nuanced analyses of the approximation error.

\subsection{Smoothness properties of the conditional velocity field}

We now investigate the estimation error from a conditional-distribution perspective in more general settings. Specifically, we do not impose an explicit structural assumption on the relationship between the covariates $X$ and the response $Y$. Instead, we place generic conditions on the conditional density of $Y$ given $X$, which are key to establishing regularity properties of the conditional velocity field. In Proposition \ref{prop: warm_up_continuity}, the Lipschitz continuity of the proxy function $g_0$ is derived as a consequence of the Gaussian setting, and approximation theory for neural networks on Lipschitz functions is then used to control the estimation error. The Lipschitz property is also central to the existence and uniqueness of solutions to initial value problems (IVPs) \citep{bris2008existence, teschl2012ordinary}. Accordingly, in this section we still expect that the proxy function $g_0$ is Lipschitz continuous in a broad class of settings.

We continue to adopt the linear interpolant $\mathcal{I}_{\mathrm{lin}}(y_0, y_1, t)=\alpha_t y_0+\beta_t y_1$ for its simplicity and wide use. The functions $\alpha_t$ and $\beta_t$ are continuously differentiable on $t\in[0, 1]$ and satisfy the boundary conditions $\alpha_0=\beta_1=1$ and $\alpha_1=\beta_0=0$. The corresponding conditional velocity field is
$$
b_0(x, y, t)
=\mathbb{E}\bigl(\dot{\alpha}_t\eta+\dot{\beta}_tY \,\big|\, Y_t=y, X=x\bigr).
$$
Under the SDR structure $Y\perp\!\!\!\perp X|R_0(X)$, Lemma \ref{lem: low_dim_veclocity} implies the existence of a function $g_0$ such that $b_0(x, y, t)\equiv g_0(R_0(x), y, t)$, provided that $\alpha_t^2+\beta_t^2\neq 0$ for all $t\in[0, 1]$. The smoothness of $g_0(z, y, t)$, with $z=R_0(x)$, is crucial for the subsequent analysis. Lemmas \ref{lem: vel_gradient_for_y}, \ref{lem: vel_gradient_for_t}, and \ref{lem: vel_gradient_for_z} characterize the gradients of $g_0$.

\begin{lemma}\label{lem: vel_gradient_for_y}
Assume that
\begin{enumerate}[label=(\roman*)]
    \item $\alpha_t\beta_t>0$ for all $t\in (0, 1)$;
    \item $\mathbb{E}[Y|R_0(X)=z]$ exists for all $z\in \mathrm{supp}(R_0(X))$.
\end{enumerate}
Then, we have
$$
\nabla_yg_0(z, y, t)=\frac{\beta_t}{\alpha_t^3}(\alpha_t\dot{\beta}_t-\dot{\alpha}_t\beta_t)\mathrm{Cov}[Y|Y_t=y, R_0(X)=z]+\frac{\dot{\alpha}_t}{\alpha_t}I_{d_y},
$$
for any $z\in \mathrm{supp}(R_0(X))$, $y\in\mathbb{R}^{d_y}$, and $t\in (0, 1)$.
\end{lemma}

\begin{lemma}\label{lem: vel_gradient_for_t}
Assume that
\begin{enumerate}[label=(\roman*)]
    \item $\alpha_t\beta_t>0$ for all $t\in (0, 1)$;
    \item $\alpha_t$ and $\beta_t$ are twice continuously differentiable on $t\in [0, 1]$.
\end{enumerate}
Then, we have
$$
\begin{aligned}
\partial_tg_0(z, y, t) =& \left[\frac{1}{\alpha_t}(\alpha_t\ddot{\beta}_t-\ddot{\alpha}_t\beta_t)-\frac{\dot{\alpha}_t}{\alpha_t^2}(\alpha_t\dot{\beta}_t-\dot{\alpha}_t\beta_t)\right]\mathbb{E}[Y|Y_t=y, R_0(X)=z] \\
& +\left(\frac{\ddot{\alpha}_t}{\alpha_t}-\frac{\dot{\alpha}_t^2}{\alpha_t^2}\right)y-\frac{\beta_t}{\alpha_t^4}(\alpha_t\dot{\beta}_t-\dot{\alpha}_t\beta_t)^2\mathrm{Cov}[\|Y\|_2^2, Y|Y_t=y, R_0(X)=z] \\
& +\frac{1}{\alpha_t^4}(\alpha_t\dot{\beta}_t-\dot{\alpha}_t\beta_t)(\alpha_t\dot{\beta}_t-2\dot{\alpha}_t\beta_t)\mathrm{Cov}[Y|Y_t=y, R_0(X)=z]\cdot y,
\end{aligned}
$$
for any $z\in \mathrm{supp}(R_0(X))$, $Y\in\mathbb{R}^{d_y}$ and $t\in (0, 1)$.
\end{lemma}

\begin{lemma}\label{lem: vel_gradient_for_z}
Let $U(\xi, z)=-\log p_1(\xi|z)$ for any $\xi\in \mathrm{supp}(Y|R_0(X)=z)$ and $z\in \mathrm{supp}(R_0(X))$. Assume that
\begin{enumerate}[label=(\roman*)]
    \item $\alpha_t\beta_t>0$ for all $t\in (0, 1)$;
    \item $\mathbb{E}[Y|R_0(X)=z]$ exists for all $z\in \mathrm{supp}(R_0(X))$;
    \item $\mathrm{supp}(Y|R_0(X)=z)\equiv\mathrm{supp}(Y)\subseteq \mathbb{R}^{d_y}$ for all $z\in \mathrm{supp}(R_0(X))$;
    \item $U(\xi, z)$ is differentiable with respect to $z$ for all $z\in \mathrm{supp}(R_0(X))$;
    \item there exists an integrable function $h^*(\xi): \mathbb{R}^{d_y}\to\mathbb{R}$ such that $\|\nabla_zU(\xi, z)p_1(\xi|z)\|_2\le h^*(\xi)$ and $\|\xi\cdot [\nabla_zU(\xi, z)]^{\top}p_1(\xi|z)\|_2\le h^*(\xi)$ for any $(\xi, z)\in \mathrm{supp}(Y, R_0(X))$.
\end{enumerate}
Then, we have
$$
\nabla_zg_0(z, y, t)=-\frac{\alpha_t\dot{\beta}_t-\dot{\alpha}_t\beta_t}{\alpha_t}\mathrm{Cov}\{Y, [\nabla_zU(Y, z)]^{\top}|Y_t=y, R_0(X)=z\},
$$
for any $z\in \mathrm{supp}(R_0(X))$, $Y\in\mathbb{R}^{d_y}$, and $t\in (0, 1)$.
\end{lemma}

Lemma \ref{lem: vel_gradient_for_y} shows that the gradient of $g_0$ with respect to $y$ is directly governed by $\alpha_t$. Due to the presence of $\alpha_t^{-1}\dot{\alpha}_t I_{d_y}$ in this gradient, it would likely diverge as $\alpha_t\to 0$ when $t\to 1$, revealing a potential singularity of $g_0$ at $t=1$. As a consequence, accurately estimating $g_0$ becomes challenging near $t=1$. A similar phenomenon arises in Lemma \ref{lem: vel_gradient_for_t}, motivating the utilization of an early-stopping strategy. In addition, Lemma \ref{lem: vel_gradient_for_z} provides an explicit expression for the gradient of $g_0$ with respect to $z$ and clarifies its dependence on the potential $U$. Conditions \textit{(ii)} in Lemma \ref{lem: vel_gradient_for_y} and \textit{(ii)}–\textit{(iv)} in Lemma \ref{lem: vel_gradient_for_z} are imposed to justify the use of the Leibniz integral rule, which allows gradients to pass under the integral.

Notably, the gradients of $g_0$ are expressed in terms of conditional expectations and covariance matrices given $Y_t$ and $R_0(X)$; the existence and boundedness of these objects are important to our analysis. We assume that the conditional density of $Y$ given $R_0(X)$ satisfies log-concavity and log-convexity properties, as stated in condition \textit{(iii)} of Lemma \ref{lem: vel_moment_bounds}. This assumption enables us to bound from above the conditional covariance matrix of $Y$ given $Y_t$ and $R_0(X)$ via the Brascamp–Lieb inequality (\cite{brascamp1976extensions}, Theorem 4.1). Condition \textit{(ii)} in Lemma \ref{lem: vel_moment_bounds} ensures that the conditional density of $Y_t$ given $R_0(X)$, evaluated at $0$, is bounded below by a positive constant, while conditions \textit{(iv)} and \textit{(v)} are included for technical convenience.

\begin{lemma}\label{lem: vel_moment_bounds}
Let $U(\xi, z)=-\log p_1(\xi|z)$ for any $\xi\in \mathrm{supp}(Y|R_0(X)=z)$ and $z\in \mathrm{supp}(R_0(X))$. Assume that
\begin{enumerate}[label=(\roman*)]
    \item $\alpha_t\beta_t>0$ for all $t\in (0, 1)$;
    \item there exists a constant $\delta>0$ such that $\|\xi\|_2\le \delta$ implies $\underline{c}\le p_1(\xi|z)\le \bar{c}$ for some constants $\underline{c}>0$ and $\bar{c}>0$, uniformly for all $z\in\mathrm{supp}(R_0(X))$;
    \item $\underline{\kappa}I_{d_y}\preceq \nabla_{\xi}^2U(\xi, z)\preceq \bar{\kappa}I_{d_y}$ for some constants $\underline{\kappa}>0$ and $\bar{\kappa}>0$, uniformly for all $\xi\in \mathrm{supp}(Y|R_0(X)=z)$ and $z\in \mathrm{supp}(R_0(X))$;
    \item $U(\xi, z)$ is continuous at $(\xi, z)$ for any $(\xi, z)\in \mathrm{supp}(Y, R_0(X))$;
    \item $\mathrm{supp}(Y|R_0(X)=z)\equiv\mathrm{supp}(Y)\subseteq \mathbb{R}^{d_y}$ is convex and close for all $z\in \mathrm{supp}(R_0(X))$, and $\mathrm{supp}(R_0(X))\subseteq \mathbb{R}^d$ is compact.
\end{enumerate}
Then, there is a constant $c^*$, which does not depend on $z, y, t$, such that
$$
\begin{aligned}
\|\mathrm{Cov}[Y|Y_t=y, R_0(X)=z]\|_2 &\le c^*\alpha_t^2, \\
\|\mathbb{E}[Y|Y_t=y, R_0(X)=z]\|_2 &\le c^*(\|y\|_2\vee 1), \\
\|\mathrm{Cov}[\|Y\|_2^2, Y|Y_t=y, R_0(X)=z]\|_2 &\le c^*\alpha_t^2(\|y\|_2\vee 1),
\end{aligned}
$$
for any $z\in \mathrm{supp}(R_0(X))$, $y\in \mathbb{R}^{d_y}$, and $t\in (0, 1)$.
\end{lemma}

Given Lemma \ref{lem: vel_moment_bounds}, Proposition \ref{prop: vel_continuity_ty} establishes the regularity of the gradients of $g_0$ with respect to the spatial variable $y$ and time $t$. The upper bounds are explicitly controlled by the decay rate of $\alpha_t$ on each compact subset of $y$. Proposition \ref{prop: vel_continuity_ty} further yields Corollary \ref{cor: vel_continuity_z}, which provides an upper bound for the norm of the gradient with respect to $z$ that depends only on the region of $y$, under suitable conditions. For concreteness, we impose a linear or quadratic growth condition of the form $\|\nabla_z U(\xi, z)\|_2\le \tilde{c}(\|\xi\|_2^{\jmath}+1)$ for $\jmath\in\{1, 2\}$; other similar growth conditions could also be considered to derive a neat upper bound. The proofs of Lemma \ref{lem: vel_moment_bounds}, Proposition \ref{prop: vel_continuity_ty}, and Corollary \ref{cor: vel_continuity_z} are provided in Supplementary Material Section B.3.

\begin{proposition}\label{prop: vel_continuity_ty}
Let $U(\xi, z)=-\log p_1(\xi|z)$ for any $\xi\in \mathrm{supp}(Y|R_0(X)=z)$ and $z\in \mathrm{supp}(R_0(X))$. Assume that
\begin{enumerate}[label=(\roman*)]
    \item $\alpha_t\beta_t>0$ for all $t\in (0, 1)$;
    \item $\alpha_t$ and $\beta_t$ are twice continuously differentiable on $t\in [0, 1]$;
    \item there exists a constant $\delta>0$ such that $\|\xi\|_2\le \delta$ implies $\underline{c}\le p_1(\xi|z)\le \bar{c}$ for some constants $\underline{c}>0$ and $\bar{c}>0$, uniformly for all $z\in\mathrm{supp}(R_0(X))$;
    \item $\underline{\kappa}I_{d_y}\preceq \nabla_{\xi}^2U(\xi, z)\preceq \bar{\kappa}I_{d_y}$ for some constants $\underline{\kappa}>0$ and $\bar{\kappa}>0$, uniformly for all $\xi\in \mathrm{supp}(Y|R_0(X)=z)$ and $z\in \mathrm{supp}(R_0(X))$;
    \item $U(\xi, z)$ is continuous at $(\xi, z)$ for any $(\xi, z)\in \mathrm{supp}(Y, R_0(X))$;
    \item $\mathrm{supp}(Y|R_0(X)=z)\equiv\mathrm{supp}(Y)\subseteq \mathbb{R}^{d_y}$ is convex and close for all $z\in \mathrm{supp}(R_0(X))$, and $\mathrm{supp}(R_0(X))\subseteq \mathbb{R}^d$ is compact.
\end{enumerate}
For any $\tau\in (0, 1)$ and $A\ge 1$, let $\psi(\tau)=\max_{t\in [0, 1-\tau]}\alpha_t^{-1}$, and
$$
\begin{aligned}
\Omega_{A, \tau} =& \{z\in\mathrm{supp}(R_0(X)), \|y\|_2\le A, t\in (0, 1-\tau)\} \\
\Omega_{\tau} =& \{z\in\mathrm{supp}(R_0(X)), y\in\mathbb{R}^{d_y}, t\in (0, 1-\tau)\}.
\end{aligned}
$$
Then, there is a constant $c^*$, which does not depend on $A$ or $t$, such that
$$
\begin{aligned}
\sup_{(z, y, t)\in \Omega_{\tau}}\|\nabla_yg_0(z, y, t)\|_2 \le & c^*\psi(\tau), \\
\sup_{(z, y, t)\in \Omega_{A, \tau}}\|\partial_tg_0(z, y, t)\|_2 \le & c^*A\psi(\tau)^2, \\
\sup_{(z, y, t)\in \Omega_{A, \tau}}\|g_0(z, y, t)\|_2 \le & c^*A\psi(\tau). \\
\end{aligned}
$$
\end{proposition}

\begin{corollary}\label{cor: vel_continuity_z}
Under the conditions in Proposition \ref{prop: vel_continuity_ty}, additionally assume that
\begin{enumerate}[label=(\roman*), start=7]
    \item $U(\xi, z)$ is differentiable with respect to $z$ for all $z\in \mathrm{supp}(R_0(X))$;
    \item there exists an integrable function $h^*(\xi): \mathbb{R}^{d_y}\to\mathbb{R}$ such that $\|\nabla_zU(\xi, z)p_1(\xi|z)\|_2\le h^*(\xi)$ and $\|\xi\cdot [\nabla_zU(\xi, z)]^{\top}p_1(\xi|z)\|_2\le h^*(\xi)$ for any $(\xi, z)\in \mathrm{supp}(Y, R_0(X))$.
\end{enumerate}
Then, there is a constant $c^{**}$ such that
$$
\|\nabla_zg_0(z, y, t)\|_2\le c^{**}\left(\mathbb{E}\left\{\left\|[\nabla_zU(Y, z)]\right\|_2^2|Y_t=y, R_0(X)=z\right\}\right)^{1/2},
$$
for any $z\in \mathrm{supp}(R_0(X))$, $y\in\mathbb{R}^{d_y}$ and $t\in (0, 1)$. Furthermore, if there exists a constant $\tilde{c}$ such that $\|\nabla_zU(\xi, z)\|_2\le \tilde{c}(\|\xi\|_2^{\jmath}+1)$ for any $(\xi, z)\in \mathrm{supp}(Y, R_0(X))$ and $\jmath=1, 2$, then for any $A\ge 1$, $\|y\|_2\le A$ implies
$$
\|\nabla_zg_0(z, y, t)\|_2\le c^{***}A^{\jmath},
$$
for any $z\in \mathrm{supp}(R_0(X))$ and $t\in (0, 1)$, where $c^{***}$ is a constant which does not depend on $A$.
\end{corollary}

We now have adequate information on the smoothness of the conditional velocity field through the gradient behavior of $g_0$. Next, we construct empirical estimators for $g_0$ and $R_0$ and analyze their sample properties.

\subsection{Empirical consistency for conditional distribution}

Recall that $R_0:\mathbb{R}^{d_x}\to\mathbb{R}^d$ and assume that $d$ is known. To empirically estimate $g_0$ and $R_0$, we independently generate random vectors $\eta_1, \dots, \eta_n$ from the standard Gaussian distribution on $\mathbb{R}^{d_y}$ and random variables $t_1, \dots, t_n$ from the uniform distribution on $[0, 1-\tau_n]$, where $\tau_n\in(0, 1/2)$ is a nonrandom quantity depending on the sample size $n$ and encodes early stopping. Given the dataset $\{(X_1, Y_1), \dots, (X_n, Y_n)\}$, we define the empirical estimators by
$$
\hat{g}, \hat{R}
=\mathop{\mathrm{argmin}}_{g\in\mathcal{F}_1, R\in\mathcal{F}_2}
\frac{1}{n}\sum_{i=1}^n
\bigl\|
\dot{\alpha}_{t_i}\eta_i+\dot{\beta}_{t_i}Y_i
-
g\bigl(R(X_i), Y_{i,t_i}, t_i\bigr)
\bigr\|_2^2,
$$
where $\mathcal{F}_1$ and $\mathcal{F}_2$ are function classes, and $Y_{i,t_i}=\alpha_{t_i}\eta_i+\beta_{t_i}Y_i$ for $i=1, \dots, n$.

Analogous to the warm-up case, we define two ODEs corresponding to continuous and discretized flows. Given the estimators $\hat{g}$ and $\hat{R}$, the ODE for the continuous flow is
$$
\mathrm{d}\tilde{Y}_t
=\hat{g}\bigl(\hat{R}(x), \tilde{Y}_t, t\bigr)\,\mathrm{d}t,
\quad
\tilde{Y}_0\sim \eta,
$$
for $t\in[0, 1-\tau_n]$ and any fixed covariate $X=x$. Incorporating time discretization, we consider
$$
\mathrm{d}\hat{Y}_t
=\hat{g}\bigl(\hat{R}(x), \hat{Y}_{\bar{t}_k}, \bar{t}_k\bigr)\,\mathrm{d}t,
\quad
t\in[\bar{t}_k, \bar{t}_{k+1}],\; k=0, \dots, K-1,
\quad
\hat{Y}_0\sim \eta,
$$
for $t\in[0, 1-\tau_n]$ and given $X=x$, where $\{\bar{t}_0, \dots, \bar{t}_K\}$ is a pre-specified time grid with $\bar{t}_0=0$ and $\bar{t}_K=1-\tau_n$. Denote by $\pi_t(x)$ the conditional distribution of $Y_t=\mathcal{I}_{\mathrm{lin}}(\eta, Y, t)$ given $X=x$. Theorem \ref{thm: general_estimation_error} provides an upper bound on the resulting estimation error from the perspective of conditional distributions.

\begin{theorem}\label{thm: general_estimation_error}
Let $U(\xi, z)=-\log p_1(\xi|z)$ for any $\xi\in \mathrm{supp}(Y|R_0(X)=z)$ and $z\in \mathrm{supp}(R_0(X))$. In addition to conditions (i)-(vi) of Proposition \ref{prop: vel_continuity_ty}, assume that
\begin{enumerate}[label=(\roman*)]
    \setcounter{enumi}{6}
    \item there exists a constant $C_1$ such that $\sup_{z\in\mathrm{supp}(R_0(X)), \|y\|_2\le A, t\in (0, 1)}\|\nabla_zg_0(z, y, t)\|_2\le C_1A$ for any $A\ge 1$;
    \item there exists a constant $C_2$ such that $\|R_0(X)\|_{\infty}< C_2$, $e_j^{\top}R_0\in\mathcal{W}_{\mathrm{Loc}}^{1, \infty}(\mathbb{R}^{d_x}, C_2)$ where $e_j$ represents the one-hot $d$-dimensional vector with its $j$-th component being 1 and others being 0, for all $j\in \{1, \dots, d\}$, and $\|X\|_{\infty}$ follows a sub-Gaussian distribution;
    \item the ODE induced by true velocity field $b_0(x, y, t)$ with Gaussian initialization admits a unique solution for any $x\in\mathrm{supp}(X)$, such that the solution at any time $t\in[0, 1]$ follows the distribution of $Y_t|X=x$.
\end{enumerate}
Define $\psi(\tau)=\max_{t\in [0, 1-\tau]}\alpha_t^{-1}$ and $\bar{\psi}(\tau)=\alpha_{1-\tau}\vee |1-\beta_{1-\tau}|$ for $\tau\in (0, 1/2)$. Let $\mathcal{F}_1=\mathcal{F}_{1, \mathrm{NN}}\cap \mathcal{F}_{\mathrm{Lip}, \Lambda_n}$. Here, $\mathcal{F}_{1, \mathrm{NN}}$ a neural network function class with an input dimensionality $d+d_y+1$, output dimensionality $d_y$, depth $L_1$, width $M_1$, uniform upper bound $\bar{\delta}_{1, n}=\psi(\tau_n)(\log n)^{1/2}\log\log n$, uniform lower bound $\underline{\delta}_{1, n}=-\psi(\tau_n)(\log n)^{1/2}\log\log n$, and $\Lambda_n=\psi(\tau_n)^2(\log n)^{1/2}\log\log n$. Let $\mathcal{F}_2=\mathcal{F}_{2, \mathrm{NN}}$ be a neural network function class with an input dimensionality $d_x$, output dimensionality $d$, depth $L_2$, width $M_2$, uniform upper bound $\bar{\delta}_{2, n}=\log\log n$, uniform lower bound $\underline{\delta}_{2, n}=-\log\log n$. Let $\Delta=\max(\bar{t}_1-\bar{t}_0, \dots, \bar{t}_K-\bar{t}_{K-1})\le n^{-1/(d^*+2)}\exp\left\{-\psi(\tau_n)^2(\log n)\right\}$. Then, for any $n\ge 3$, we have
$$
\begin{aligned}
& \mathbb{E}\left[W_2(\pi_1(X), \hat{\pi}_{1-\tau_n}(X))\right] \\
\le & c_7\left(\psi(\tau_n)^2n^{-\frac{1}{d^*+2}}(\log n)^{5/2}[\log (n\vee \psi(\tau_n))]^{1/2}\exp\left\{\psi(\tau_n)^2(\log n)^{1/2}\log\log n\right\}+\bar{\psi}(\tau_n)\right),
\end{aligned}
$$
where $d^*=\max(d_x, d+d_y+1)$ and $c^*$ is a constant which does not depend on $n$.
\end{theorem}

Theorem \ref{thm: general_estimation_error} explicitly reveals how the interpolation scheme, via $\alpha_t$ and $\beta_t$, affects the upper bound of the estimation error. When $\tau\to 0^+$, the function $\psi(\tau)$ diverges whereas $\bar{\psi}(\tau)$ converges to $0$. By carefully balancing $\psi(\tau_n)$ and $\bar{\psi}(\tau_n)$, we obtain a vanishing estimation error (see, for example, Corollary \ref{cor: estimation_error_conv_example}; we omit its proof since it is a straightforward application of Theorem \ref{thm: general_estimation_error}).

Theorem \ref{thm: general_estimation_error} thus showcases that the empirical GenSDR estimator of $R_0$ achieves a convergent estimation error at the level of conditional distributions, corroborating its ability to recover the information encoded in the central $\sigma$-field. To the best of our knowledge, such a theoretical guarantee is unavailable for existing methods such as GMDDNet or BENN. We remark that conditions \textit{(vii)} and \textit{(viii)} are introduced for technical convenience. Moreover, the upper bound in Theorem \ref{thm: general_estimation_error} can be sharpened by developing a more nuanced approximation theory for the conditional velocity field that separates different components of Lipschitz regularity, which however lies beyond the scope of this work.

\begin{corollary}\label{cor: estimation_error_conv_example}
Let $\alpha_t=1-t$, $\beta_t=t$, and $\tau_n=(\log n)^{-1/4}\log\log n$. Then, under the conditions of Theorem \ref{thm: general_estimation_error}, it follows that $\mathbb{E}[W_2(\pi_1(X), \hat{\pi}_{1-\tau_n}(X))]\to 0$ as $n\to \infty$.
\end{corollary}

\section{Beyond Euclidean responses}\label{sec: beyond_euclidean}

With the continuing development of data collection technologies, non-Euclidean data objects arise in numerous statistical applications. These complex data types, which include but are not limited to symmetric positive definite (SPD) matrices, graph Laplacians of networks, and probability density functions, pose substantial challenges for statistical analysis \citep{petersen2019frechet, qiu2024random, bhattacharjee2025nonlinear}. In the realm of nonlinear SDR, most existing work focuses on Euclidean responses, leaving non-Euclidean scenarios largely unexplored. This section extends our GenSDR method to accommodate non-Euclidean responses.

Unlike Euclidean objects, non-Euclidean ones possess intrinsic geometric structure, which is the central difficulty to address. In Sections \ref{sec: illustrative_example} and \ref{sec: empirical_consistency}, we employ a linear interpolant. However, such an interpolation scheme may become meaningless under non-Euclidean geometry, since intermediate states are not guaranteed to lie in the underlying response space. It is therefore necessary to adapt the interpolation mechanism to the specific geometry. Broadly, there are two strategies. The first constructs a geometry-aware interpolant that complies with a stay-within-space constraint; see \cite{huang2022riemannian} and \cite{chen2024flow} for practical examples. The second maps non-Euclidean objects into a Euclidean space while attempting to preserve all relevant information. In this section, we adopt the latter approach, which admits a more transparent theoretical underpinning. Specifically, we build on \cite{zhang2024dimension}, which develops linear SDR for metric space–valued responses. By combining the ensemble technique introduced in \cite{yin2011sufficient} and further developed in \cite{zhang2024dimension} with our GenSDR method originally proposed for Euclidean responses, we obtain a novel procedure for nonlinear SDR with non-Euclidean responses.

Let $\Omega_Y$ denote the (possibly non-Euclidean) domain of the response $Y$. A function class $\mathcal{H}$ consisting of mappings from $\Omega_Y$ to $\mathbb{R}$ is called an eSDR-ensemble (exhaustive SDR-ensemble) if
$$
\mathcal{G}_{Y|X} \subseteq \sigma\bigl\{\mathcal{G}_{h(Y)|X} : h \in \mathcal{H}\bigr\}.
$$
Informally, an eSDR-ensemble \emph{encodes} the geometry of $\Omega_Y$ and recovers the central $\sigma$-field $\mathcal{G}_{Y|X}$ by aggregating the snapshot $\sigma$-fields induced by its elements. The following Lemma~\ref{lem: ensemble_chracter} shows that our eSDR-ensemble has the same characterization property as the CS-ensemble in \cite{zhang2024dimension} (see Lemma~1 therein). Here, $L_2(\mathbb{P}_Y)$ denotes the set of functions on $\Omega_Y$ that are square-integrable with respect to the distribution of $Y$, namely $\mathbb{P}_Y$; $\mathrm{span}(\mathcal{H}) = \{\sum_{i=1}^k a_i h_i : k \in \mathbb{N}, a_1,\dots,a_k \in \mathbb{R}, h_1,\dots,h_k \in \mathcal{H}\}$; and $\mathcal{B} = \{\mathds{1}_B : B \text{ is a Borel set in } \Omega_Y\}$ is the collection of indicator functions of Borel subsets of $\Omega_Y$. 

\begin{lemma}\label{lem: ensemble_chracter}
If $\mathcal{H}$ is a subset of $L_2(\mathbb{P}_Y)$ and $\mathrm{span}(\mathcal{H})$ is dense in $\mathcal{B}$ with respect to the $L_2(\mathbb{P}_Y)$-metric, then $\mathcal{H}$ is an eSDR-ensemble.
\end{lemma}

Given an eSDR-ensemble $\mathcal{H}$ and an arbitrary $h \in \mathcal{H}$, we can construct a stochastic interpolation model using $\mathcal{I}_{\mathrm{lin}}(\eta, h(Y), t)$, since $h(Y) \in \mathbb{R}$, where $\eta$ is a standard Gaussian random variable. Suppose that $R^{\dagger} : \mathbb{R}^{d_x} \to \mathbb{R}^q$ with $q \ge d$ is a function such that, for any $h \in \mathcal{H}$, there exists a function $g_h^{\dagger}$ satisfying
$$
g_h^{\dagger}, R^{\dagger}
=
\mathop{\mathrm{argmin}}_{g, f}
\int_0^1
\mathbb{E}
\Bigl[
\partial_t \mathcal{I}_{\mathrm{lin}}(\eta, h(Y), t)
-
g\bigl(f(X), Y_t, t\bigr)
\Bigr]^2 \,\mathrm{d}t.
$$
Then we promisingly have $\mathcal{G}_{h(Y)|X} \subseteq \sigma\bigl(R^{\dagger}(X)\bigr)$ for every $h \in \mathcal{H}$. Since $\mathcal{H}$ is an eSDR-ensemble, the arbitrariness of $h$ yields
$$
\mathcal{G}_{Y|X}
\subseteq
\sigma\bigl\{\mathcal{G}_{h(Y)|X} : h \in \mathcal{H}\bigr\}
\subseteq
\sigma\bigl(R^{\dagger}(X)\bigr).
$$
Thus, $R^{\dagger}$ is a sufficient representation for modeling non-Euclidean $Y$. This observation underpins our extension of GenSDR.

For practical implementation, we approximate $\mathcal{H}$ by a finite subset $\{h_1, \dots, h_m\} \subseteq \mathcal{H}$ with $m \in \mathbb{N}$, and estimate the ground-truth transformation $R_0$ by solving
$$
g_1^*,\dots,g_m^*,R^*
=
\mathop{\mathrm{argmin}}_{\{g_1,\dots,g_m\}\in\mathcal{F}_1^{\otimes m}, f\in\mathcal{F}_2}
\frac{1}{m}
\sum_{\ell=1}^m
\int_0^1
\mathbb{E}
\Bigl[
\partial_t \mathcal{I}_{\mathrm{lin}}(\eta, h_{\ell}(Y), t)
-
g_{\ell}\bigl(f(X), Y_{t, \ell}, t\bigr)
\Bigr]^2
\,\mathrm{d}t,
$$
where $Y_{t, \ell}=\mathcal{I}_{\mathrm{lin}}(\eta, h_{\ell}(Y), t)$, $\mathcal{F}_1^{\otimes m} = \mathcal{F}_1 \times \cdots \times \mathcal{F}_1$, and $\mathcal{F}_1, \mathcal{F}_2$ are pre-specified function classes.

At the sample level, given observations $\{(X_1, Y_1), \dots, (X_n, Y_n)\}$, we independently generate $\{\eta_1, \dots, \eta_n\}$ from the standard univariate Gaussian distribution and $\{t_1,\dots,t_n\}$ from the uniform distribution on $[0, 1-\tau_n]$, where $\tau_n$ is a small pre-specified quantity for early stopping and stability. We then obtain empirical estimators via
$$
\hat{g}_1,\dots,\hat{g}_m,\hat{R}
=
\mathop{\mathrm{argmin}}_{\{g_1,\dots,g_m\}\in\mathcal{F}_1^{\otimes m}, f\in\mathcal{F}_2}
\frac{1}{mn}
\sum_{\ell=1}^m
\sum_{i=1}^n
\left[
\partial_t \mathcal{I}_{\mathrm{lin}}(\eta_i, h_{\ell}(Y_i), t_i)
-
g_{\ell}\bigl(f(X_i), Y_{i,t_i, \ell}, t_i\bigr)
\right]^2,
$$
where $Y_{i,t_i, \ell} = \mathcal{I}_{\mathrm{lin}}(\eta_i, h_{\ell}(Y_i), t_i)$. To specify $\mathcal{H}$, we employ function families induced by bounded cc-universal kernels tailored to the task, leveraging the analogy between our eSDR-ensemble and the CS-ensemble in \cite{zhang2024dimension}.

\section{Numerical experiments}\label{sec: numerical_experiments}

\subsection{Simulation I: Euclidean responses}

This subsection demonstrates the performance of GenSDR in Euclidean response settings, compared with GSIR \citep{lee2013general}, GMDDNet \citep{chen2024deep}, DDR \citep{huang2024deep}, and BENN \citep{tang2024belted}. GenSDR was implemented in Python using the PyTorch framework. We employed lightweight neural networks for the representational transformation $R$ and the candidate proxy velocity field $g$, with architectures $d_x-64-256-128-d$ and $(d+d_y+1)-64-256-128-d_y$, respectively. Throughout, we assumed that the intrinsic dimension $d$ was known. The straight-line interpolant $\mathcal{I}_{\mathrm{sl}}(y_0, y_1, t)=(1-t)y_0+ty_1$ was utilized in all experiments. The early stopping hyperparameter was fixed as $\tau_n=0.001$, and the Adam optimizer was adopted with a learning rate of 0.001.

Implementation of GSIR was obtained from the \texttt{nsdr} package in \textsf{R}, and the implementation of DDR was adapted from the official GitHub repository. We obtained the code for GMDDNet from the authors of \cite{chen2024deep}. BENN was implemented manually following Algorithm 1 in \cite{tang2024belted}, using an ensemble family generated by the Gaussian kernel with 1000 ensemble elements to construct the loss function. In order to ensure a fair comparison, we used identical network architectures for the representational transformations in GenSDR, GMDDNet, DDR, and BENN.

We considered the following four data-generating mechanisms.
\begin{description}
    \item[(A)] $Y=h_0(X)\cdot W$, where
    $$
    \begin{aligned}
    h_0(X) &= (1+|X_{(2)}|)^{-2}\exp(X_{(1)}), \\
    W &= \bigl(W_1,\; W_2W_4+(W_3+10)(1-W_4)-\exp(W_1)\bigr)^{\top},
    \end{aligned}
    $$
    with $W_1\sim Ga(3, 5)$, $W_2\sim t(3)$, $W_3\sim t(3)$, $W_4\sim B(1/2)$ (where $B(1/2)$ denotes the Bernoulli distribution with success probability $1/2$), and $W_1, W_2, W_3, W_4$, and $X$ mutually independent;
    \item[(B)] $Y=h_0(X)+\gamma W$, where
    $$
    h_0(X)=\Bigl(X_{(1)}^2\exp(2X_{(4)}^2),\ (1+3X_{(2)}-X_{(3)})^{4/3}\mathds{1}(X_{(1)}^{1/2}>0.8)\Bigr)^{\top},
    $$
    and $W=(W_1, W_2)^{\top}$ with $W_1\sim\chi^2_3$ and $W_2\sim\chi^2_5$, where $W_1, W_2$, and $X$ are mutually independent;
    \item[(C)] $Y=h_0(X)+\mathrm{diag}\{h_1(X)\}\cdot W$, where
    $$
    \begin{aligned}
    h_0(X) &= \bigl(4\cos(\pi X_{(3)}), 1, \dots, 1\bigr)^{\top}, \\
    h_1(X) &= \Bigl(5\bigl[(X_{(1)}+\cdots+X_{(6)})/6\bigr]^3, \log(1+3X_{(2)}), 1, \dots, 1\Bigr)^{\top},
    \end{aligned}
    $$
    and $W=(W_1, \dots, W_{d_y})^{\top}$ with $W_j\overset{\mathrm{i.i.d.}}{\sim} Ga(3, 1)$ for $j=1, \dots, d_y$, and $W$ and $X$ mutually independent;
    \item[(D)] $Y=\bigl(W_1h_0(X)+W_2(1-h_0(X)),\; W_1h_1(X)+W_2(1-h_1(X))\bigr)^{\top}$, where
    $$
    \begin{aligned}
    h_0(X) &= \mathds{1}\{h_3(X)>0\}, \\
    h_1(X) &= \mathds{1}\{h_3(X)>h_4(X)\}, \\
    h_3(X) &= \max\bigl(2X_{(1)}X_{(2)}-1,\ \sin\bigl(\pi (X_{(3)}+X_{(4)})\bigr)\bigr), \\
    h_4(X) &= \min\bigl(X_{(2)}, X_{(3)}\bigr),
    \end{aligned}
    $$
    and $W_1\sim La(1, 2)$ (Laplace distribution) and $W_2\sim N(-1, 1)$, with $W_1, W_2$, and $X$ mutually independent.
\end{description}

In setting A, we examined different distributions for the covariates $X$, specifically the uniform distribution $U([0, 1]^{d_x})$, the isotropic Gaussian distribution $N(1_{d_x}, I_{d_x})$, and the anisotropic Gaussian distribution $N(1_{d_x}, HH^{\top})$, where $H=I_{d_x}-(1+d_x)^{-1}1_{d_x}1_{d_x}^{\top}$. In setting B, we varied the noise level with $\gamma\in\{0.1, 0.3, 0.5\}$. In setting C, we considered $d_y\in\{5, 10, 20\}$ to assess performance as the response dimension increased. In setting D, the sufficient representations were binary, and we considered sample sizes $n\in\{1000, 2000, 3000\}$.

In settings B, C, and D, the covariates $X$ were generated from $U([0, 1]^{d_x})$. In settings A, B, and C, we generated $n=1000$ independent training observations. For all four settings, we fixed $d_x=50$. The number of training epochs for GenSDR was set to 50, except in setting C with $d_y=20$, where it was increased to 100 to ensure algorithmic convergence.

For evaluation, we generated 1000 independent test observations per configuration and computed the associated sufficient representations. Distance correlation \citep{szekely2007measuring} was employed to quantify similarity between the true and estimated representations, with larger values indicating better performance. Each configuration was replicated 100 times. A unified random seed was set across all methods to ensure reproducibility.

\begin{table}[t]
\centering
\caption{Average distance correlations between true sufficient representations and estimated representations in Euclidean response settings, based on 100 replications. Standard deviations are reported in parentheses. The best result in each setting is highlighted in bold.}
\begin{tabular}{ccccccc}
\begin{tabular}{ccccccc}
\toprule
                             &  & GenSDR        & BENN          & DDR           & GMDDNet       & GSIR          \\ \hline
\multirow{3}{*}{Setting A}   & $U([0, 1]^d)$ & \textbf{0.954 (0.010)} & 0.857 (0.035) & 0.318 (0.059) & 0.775 (0.068) & 0.941 (0.006) \\
                             & $N(1_{d_x}, I_{d_x})$ & 0.804 (0.020) & 0.676 (0.029) & 0.451 (0.075) & 0.664 (0.054) & \textbf{0.806 (0.011)} \\
                             & $N(1_{d_x}, HH^{\top})$ & \textbf{0.810 (0.019)} & 0.689 (0.032) & 0.467 (0.081) & 0.690 (0.041) & 0.808 (0.012) \\ \hline
\multirow{3}{*}{Setting   B} & $\sigma=0.1$ & \textbf{0.881 (0.013)} & 0.833 (0.018) & 0.871 (0.022) & 0.868 (0.010) & 0.793 (0.009) \\
                             & $\sigma=0.3$ & \textbf{0.843 (0.020)} & 0.768 (0.022) & 0.768 (0.023) & 0.828 (0.025) & 0.783 (0.011) \\
                             & $\sigma=0.5$ & \textbf{0.765 (0.032)} & 0.647 (0.034) & 0.641 (0.034) & 0.761 (0.027) & 0.762 (0.014) \\ \hline
\multirow{3}{*}{Setting   C} & $d_y=5$ & \textbf{0.852 (0.020)} & 0.763 (0.034) & 0.766 (0.029) & 0.758 (0.026) & 0.784 (0.014) \\
                             & $d_y=10$ & \textbf{0.832 (0.020)} & 0.730 (0.031) & 0.773 (0.025) & 0.718 (0.035) & 0.774 (0.015) \\
                             & $d_y=20$ & \textbf{0.828 (0.020)} & 0.662 (0.044) & 0.735 (0.134) & 0.693 (0.023) & 0.753 (0.017) \\ \hline
\multirow{3}{*}{Setting   D} & $n=1000$ & \textbf{0.703 (0.041)} & 0.519 (0.046) & 0.364 (0.045) & 0.535 (0.072) & 0.683 (0.019) \\
                             & $n=2000$ & \textbf{0.761 (0.032)} & 0.559 (0.065) & 0.389 (0.042) & 0.629 (0.061) & 0.719 (0.016) \\
                             & $n=3000$ & \textbf{0.809 (0.027)} & 0.575 (0.070) & 0.414 (0.040) & 0.697 (0.045) & 0.732 (0.013) \\
\bottomrule
\end{tabular}
\end{tabular}
\label{tab: simulation_euclidean}
\end{table}

\begin{figure}[!t]
    \centering
    \includegraphics[width=0.7\linewidth]{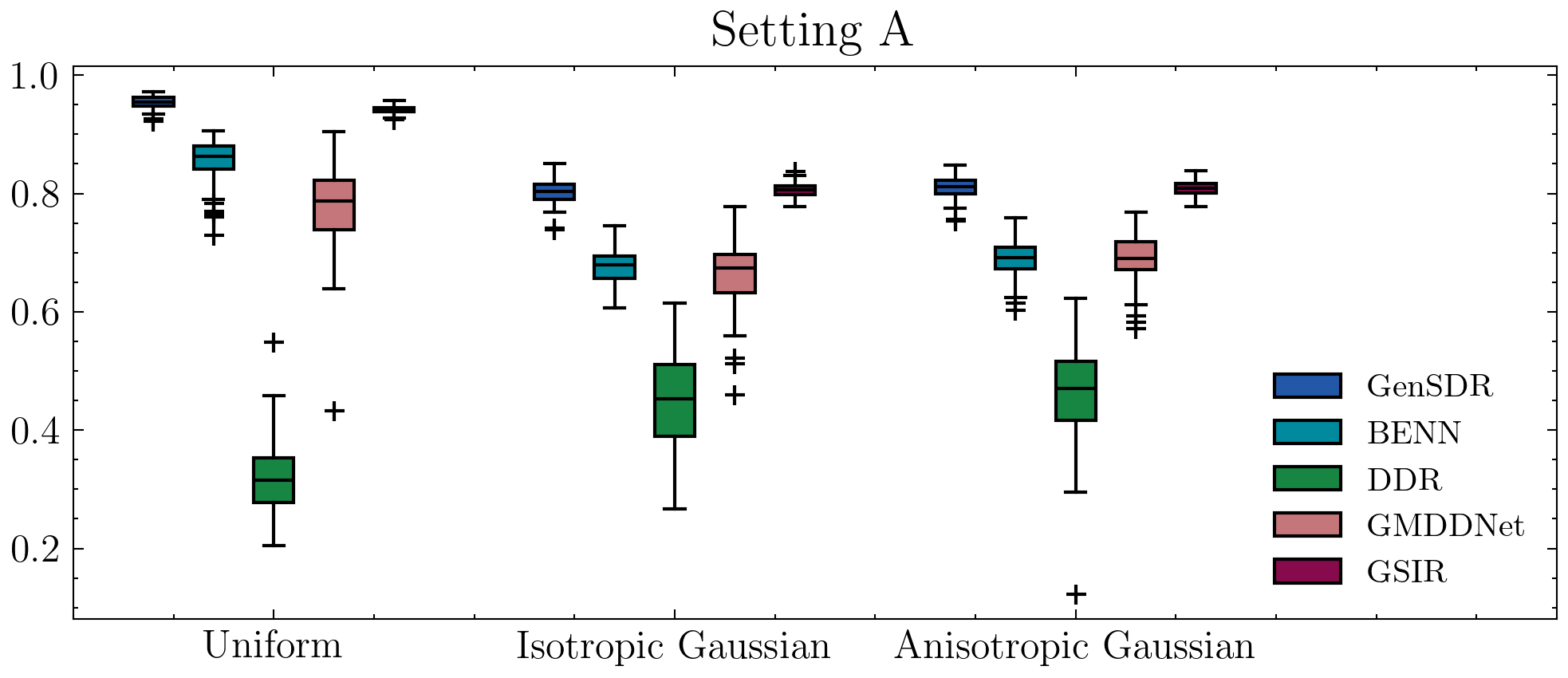}
    \includegraphics[width=0.7\linewidth]{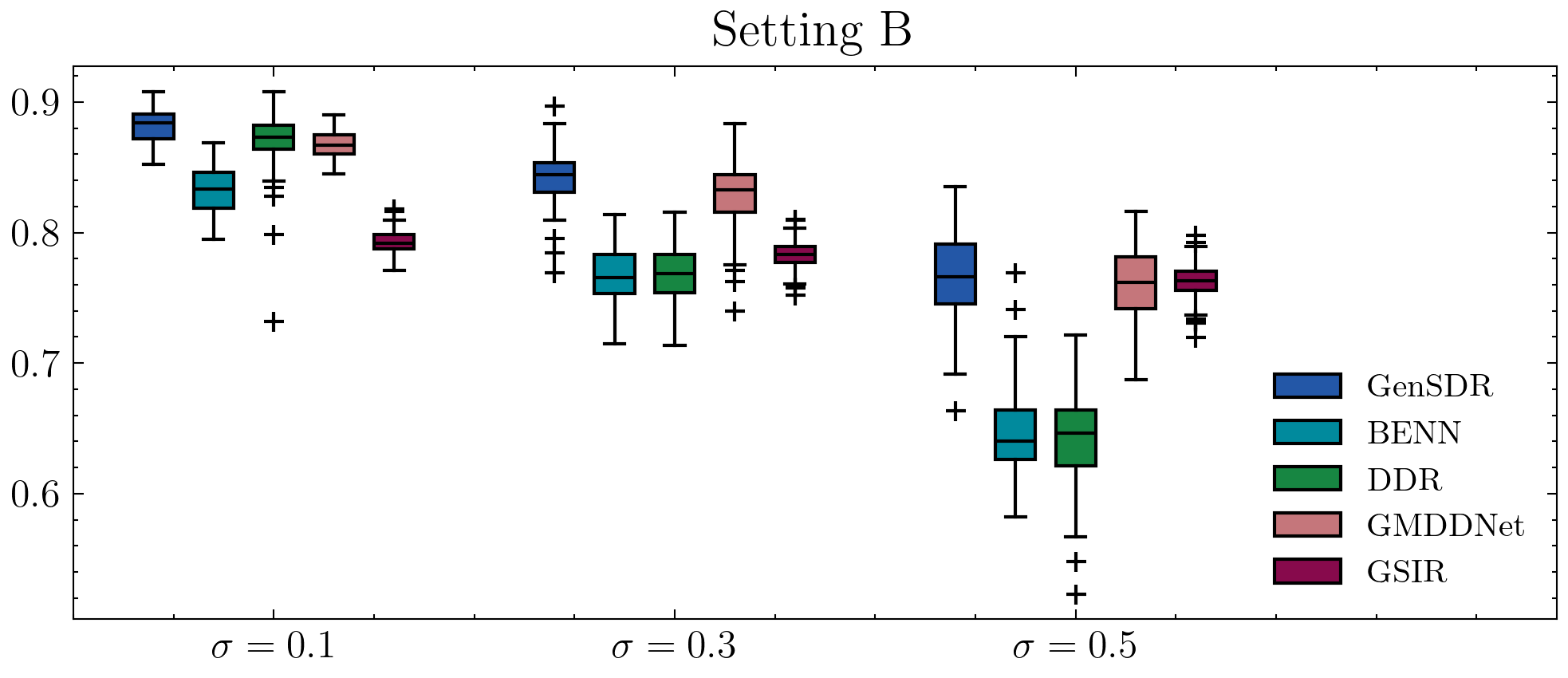}
    \includegraphics[width=0.7\linewidth]{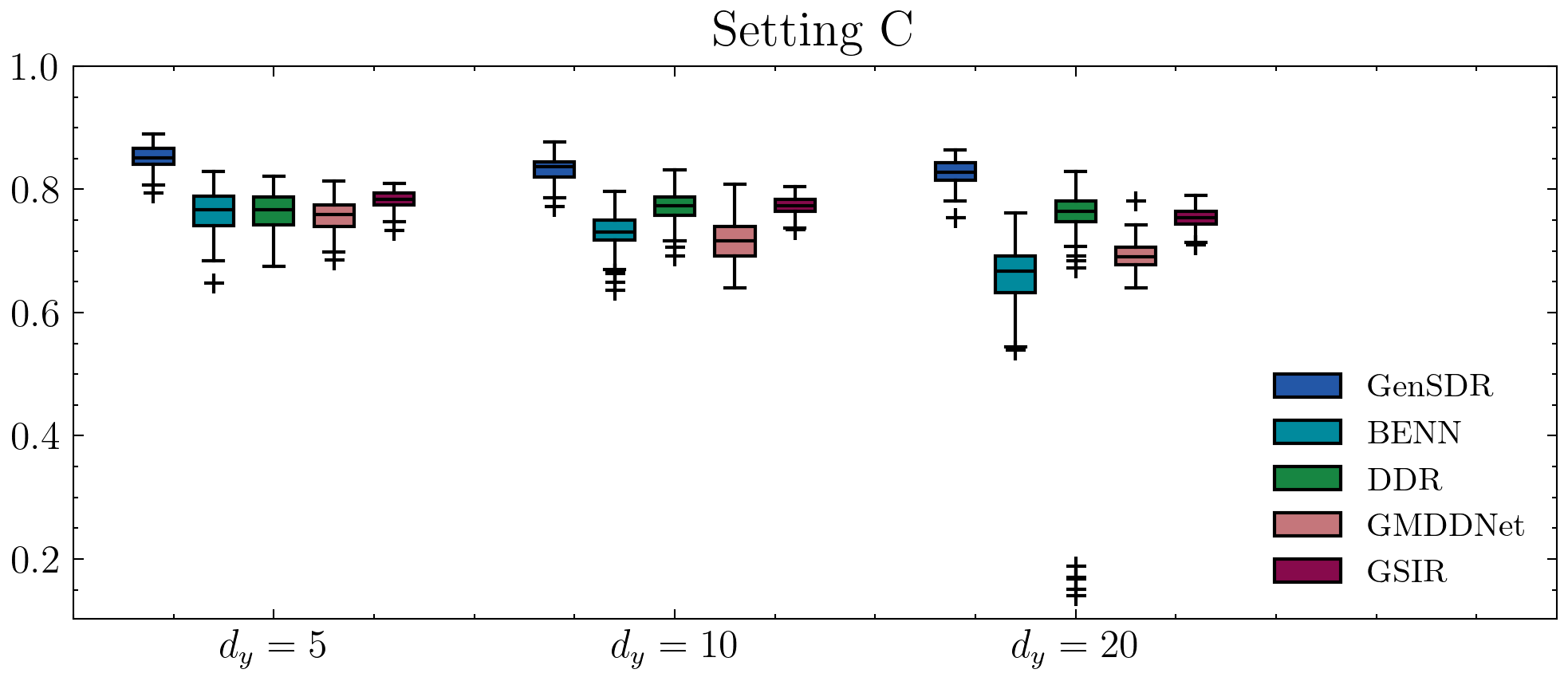}
    \includegraphics[width=0.7\linewidth]{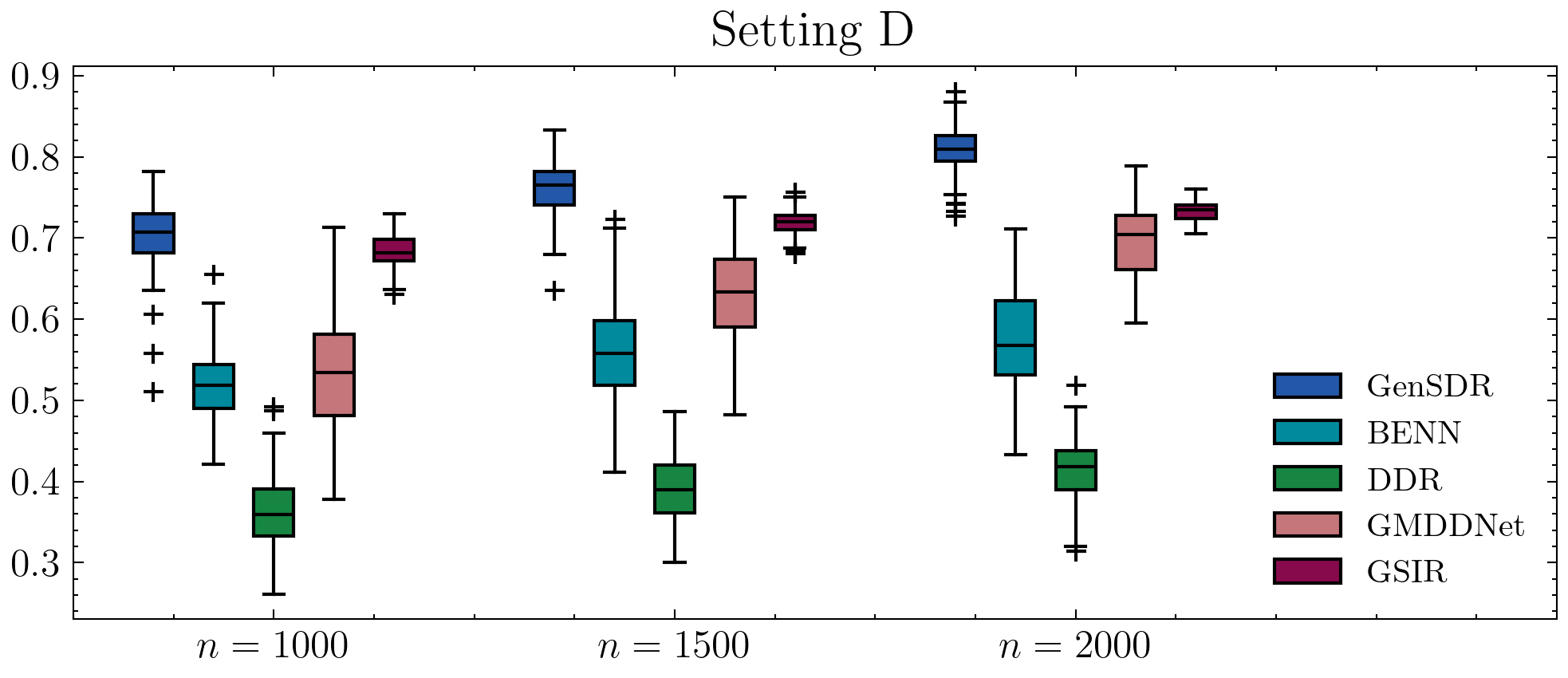}
    \caption{Boxplots of distance correlations between true sufficient representations and estimated representations in Euclidean response settings, based on 100 replications.}
    \label{fig: simulation_euclidean}
\end{figure}

From Table \ref{tab: simulation_euclidean} and Figure \ref{fig: simulation_euclidean}, GenSDR achieved the best performance in most configurations. In setting A, GenSDR and GSIR both yielded high-quality latent representations, whereas DDR struggled to recover the sufficient structures. In setting B, GenSDR and GMDDNet performed well for regression-type problems, with GenSDR demonstrating notable robustness as the noise level increased. In setting C, GenSDR effectively handled large-dimensional responses, benefiting from its additive formulation, which avoids the need for ensembles, slicing, or kernelization on $Y$. In setting D, GenSDR showcased impressive performance for binary sufficient representations, underscoring its versatility across diverse representation types.

\subsection{Simulation II: SPD responses}

We next evaluate GenSDR when the responses are symmetric positive definite (SPD) matrices. To harness the non-Euclidean geometry, we used the cc-universal kernel $\mathcal{K}(y_1, y_2)=\exp(-\omega\,\mathrm{d}_{\mathrm{F}}(y_1, y_2))$ to construct the ensemble family $\mathcal{H}$, as suggested in \cite{zhang2024dimension}, where $\mathrm{d}_{\mathrm{F}}(y_1, y_2)=\|y_1-y_2\|_{\mathrm{F}}$ and $\|\cdot\|_{\mathrm{F}}$ denotes the Frobenius norm. Specifically, we set $\mathcal{H}=\{\mathcal{K}(\cdot, y): y\in\mathcal{Y}\subseteq\Omega_Y\}$. In our experiments, $\mathcal{Y}$ (the reference set) was formed by randomly sampling one half of all observed responses. The parameter $\omega$ was chosen as the inverse median of all Frobenius distances between elements of the full set $\{Y_1, \dots, Y_n\}$ and the reference set $\mathcal{Y}$. All remaining hyperparameters of GenSDR were kept identical to those used in the Euclidean response experiments. As a benchmark, we included the Fr{\'e}chet cumulative covariance net (FCCov-Net; \cite{yuan2025fr}) method, which is proved unbiased at the $\sigma$-field level for random object responses in general metric spaces. The implementation of FCCov-Net was obtained from the authors of \cite{yuan2025fr}.

To generate SPD matrix-valued responses, we employed the symmetric matrix variate normal distribution \citep{schwartzman2006random}. An $r\times r$ random matrix $Z$ is said to follow the standard symmetric matrix variate normal distribution $N_{rr}(0, I_r)$ if it is symmetric with independent $N(0, 1)$ diagonal entries and independent $N(0, 1/2)$ off-diagonal entries. We considered the following two settings, where $\log(\cdot)$ denotes the matrix logarithm.
\begin{description}
    \item[(E)] $\log Y=0.1Z+\log D(X)$, where
    $$
    \begin{aligned}
    D(X) &=
    \begin{pmatrix}
        1 & h_0(X) \\
        h_0(X) & 1
    \end{pmatrix}, \\
    h_0(X) &= \frac{\exp(X_{(1)}+X_{(2)})-1}{\exp(X_{(1)}+X_{(2)})+1},
    \end{aligned}
    $$
    with $Z\sim N_{22}(0, I_2)$;
    \item[(F)] $\log Y=h_0(X)\cdot Z+\log D$, where
    $$
    \begin{aligned}
    D &=
    \begin{pmatrix}
        1 & -0.5 \\
        -0.5 & 1
    \end{pmatrix}, \\
    h_0(X) &= X_{(1)}X_{(2)},
    \end{aligned}
    $$
    with $Z\sim N_{22}(0, I_2)$.
\end{description}

In setting E, we varied the distribution of $X$ among $U([0, 1]^{d_x})$, $N(0, I_{d_x})$, and $N(0, HH^{\top})$, where $H=I_{d_x}-(1+d_x)^{-1}1_{d_x}1_{d_x}^{\top}$, while fixing the sample size at $n=1000$. In setting F, we considered sample sizes $n\in\{500, 1000, 1500\}$. The covariate dimension was set to $d_x=50$ in both settings. For each configuration, 1000 test observations were generated for evaluation. Analogous to the Euclidean case, distance correlation was utilized as the performance metric, and each configuration was replicated 100 times.

\begin{table}[t]
\centering
\caption{Average distance correlations between true sufficient representations and estimated representations in SPD matrix-valued response settings, based on 100 replications. Standard deviations are reported in parentheses. The best result in each setting is highlighted in bold.}
\begin{tabular}{ccc}
\begin{tabular}{cccc}
\toprule
                           &  & GenSDR        & FCCov-Net     \\ \hline
\multirow{3}{*}{Setting E} & $U([0, 1]^d)$ & \textbf{0.974 (0.013)} & 0.881 (0.011) \\
                           & $N(0, I_{d_x})$ & \textbf{0.973 (0.003)} & 0.759 (0.020) \\
                           & $N(0, HH^{\top})$ & \textbf{0.973 (0.003)} & 0.767 (0.022) \\ \hline
\multirow{3}{*}{Setting F} & $n=500$ & \textbf{0.859 (0.016)} & 0.814 (0.020) \\
                           & $n=1000$ & \textbf{0.894 (0.013)} & 0.841 (0.012) \\
                           & $n=1500$ & \textbf{0.916 (0.011)} & 0.850 (0.013) \\
\bottomrule
\end{tabular}
\end{tabular}
\label{tab: simulation_spd}
\end{table}

\begin{figure}[t]
    \centering
    \includegraphics[width=0.8\linewidth]{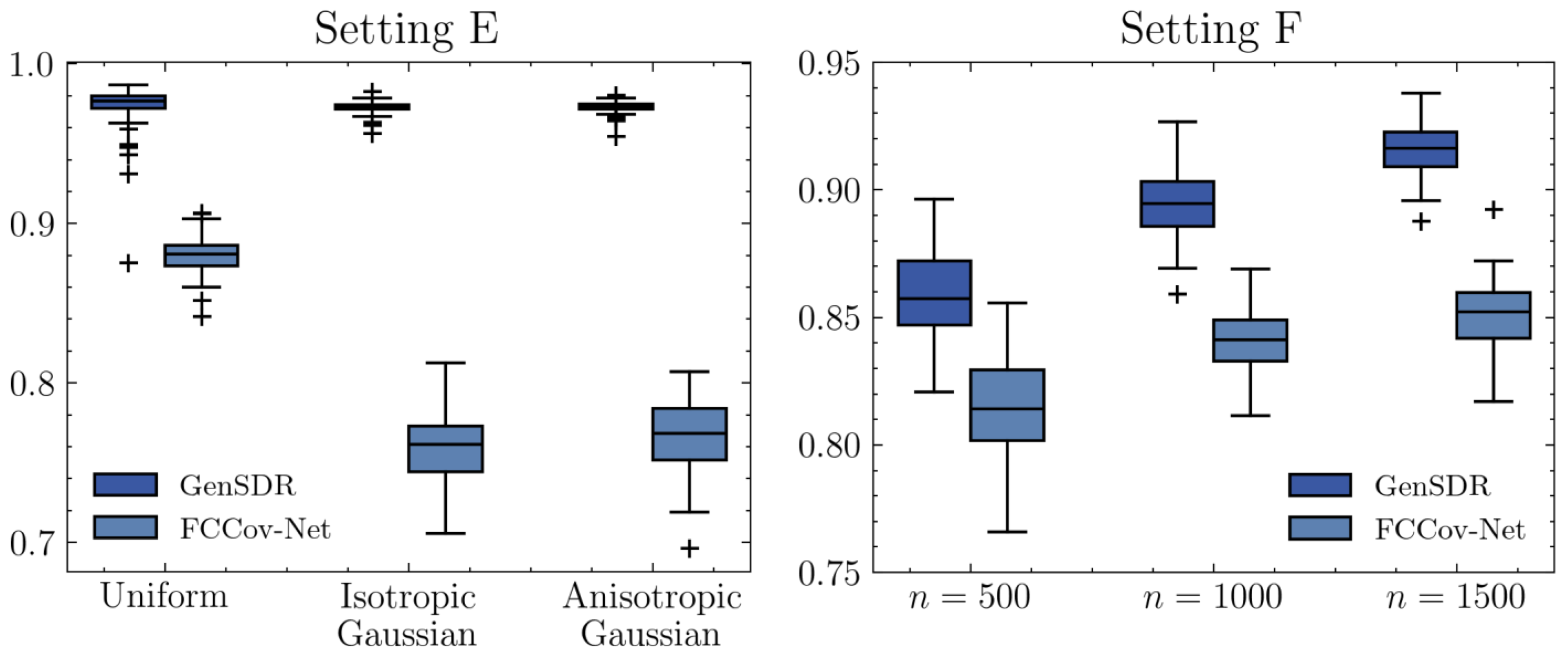}
    \caption{Boxplots of distance correlations between true sufficient representations and estimated representations in SPD matrix-valued response settings, based on 100 replications.}
    \label{fig: simulation_spd}
\end{figure}

Table \ref{tab: simulation_spd} and Figure \ref{fig: simulation_spd} show that GenSDR exhibited consistently superior performance; in particular, GenSDR outperformed FCCov-Net in all configurations. In setting E, GenSDR attained near-perfect distance correlations and displayed strong robustness to changes in the distribution of $X$. In setting F, GenSDR yielded more informative representations than FCCov-Net across all sample sizes. These results indicate that GenSDR is highly competent for SPD matrix-valued responses and is promising for broader non-Euclidean response scenarios.

\begin{figure}[t]
    \centering
    \includegraphics[width=\linewidth]{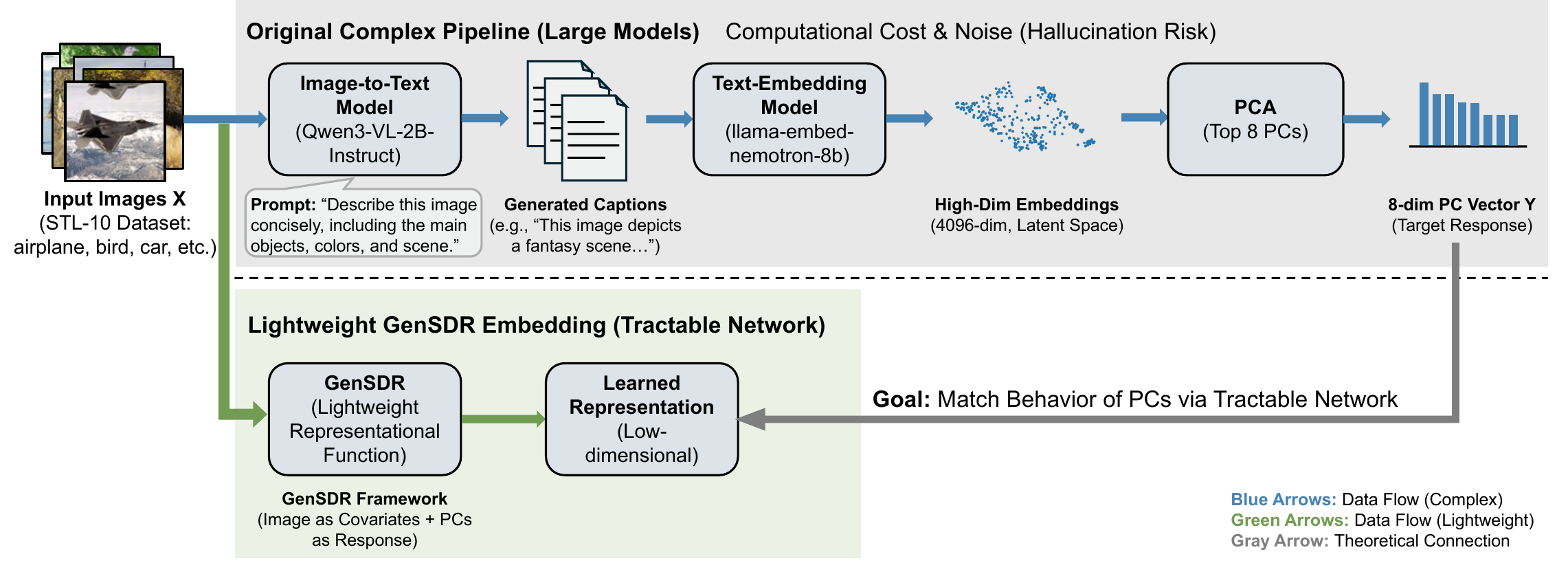}
    \caption{Illustration of the real-data analysis. We utilized GenSDR to distill a complex pipeline involving large models to a lightweight, more tractable representational network.}
    \label{fig: real_data_illustration}
\end{figure}

\subsection{Real data}

\begin{figure}[t]
    \centering
    \includegraphics[width=0.9\linewidth]{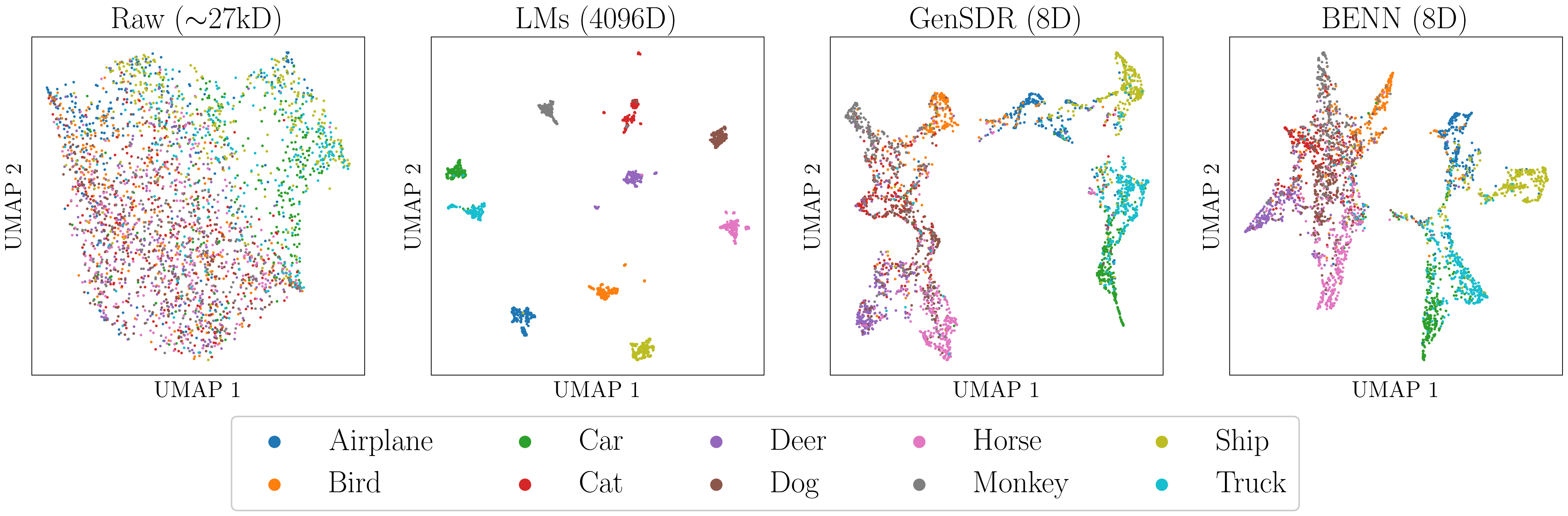}
    \caption{UMAP visualizations of raw image pixels, original embeddings from large models (LMs), the GenSDR representations, and the BENN representations on testing set.}
    \label{fig: stl_10}
\end{figure}

As an illustrative real-world application, we use GenSDR to distill a complex representation-generation pipeline into a more lightweight network. We consider the STL-10 dataset \citep{coates2011analysis}, originally designed for developing unsupervised feature learning, deep learning, and self-taught learning algorithms. The dataset contains 13,000 images from 10 categories, specifically airplane, bird, car, cat, deer, dog, horse, monkey, ship, and truck. We initially adopted uniform manifold approximation and projection (UMAP; \cite{mcinnes2018umap}) for visualization of the raw image pixels, as shown in the first panel of  Figure~\ref{fig: stl_10}. Noticeably, this direct unsupervised dimensionality reduction method failed to capture meaningful features or semantics, resulting in an unsatisfactory low-dimensional layout.

We then designed a two-stage pipeline that leveraged state-of-the-art pre-trained models for more effective information extraction from images.
First, a pre-trained image-to-text model, Qwen3-VL-2B-Instruct \citep{yang2025qwen3technicalreport, Qwen2.5-VL}, was prompted with ``Describe this image concisely, including the main objects, colors, and scene." to generate a caption for each image. An example caption is shown below.
\begin{quote}
\textit{This image depicts a fantasy scene featuring a rider on a golden, ornate horse. The horse is adorned with intricate, glowing green and gold patterns, and the rider is dressed in a red and gold outfit. The setting appears to be a dimly lit, ancient stone structure, possibly a temple or a castle, with a dark, atmospheric background.}
\end{quote}
Second, we applied a pre-trained text-embedding model, llama-embed-nemotron-8b \citep{babakhin2025llamaembednemotron8buniversaltextembedding}, to obtain a 4096-dimensional embedding for each caption. From the UMAP visualization of the 4096-dimensional embeddings, shown in the second panel of Figure~\ref{fig: stl_10}, we observe that the large models further facilitated mapping the original images into a semantic latent space, rendering the formation of meaningful clusters.
However, this procedure is computationally expensive and relies on large language models that may hallucinate \citep{huang2025survey}, thereby introducing spurious or misleading semantic information, which we regard as noise. By leveraging nonlinear SDR, we aim to recover low-dimensional representations through a substantially more tractable representational function while remaining faithful to the original images.

Among the 13,000 images with associated embeddings, we randomly sampled 10,000 for training and used the remaining 3,000 for testing. Given the relatively modest sample size and the high embedding dimension, we performed principal component analysis on the training embeddings and retained the top 8 principal components (PCs) as the response, preserving approximately 40\% of the total variance. Accordingly, the covariates $X$ were the raw image pixels, and the response $Y$ was the 8-dimensional PC vector. Our goal is to use GenSDR to learn a representational function that behaves comparably to the PCs derived from large pre-trained models, but implemented via a smaller and more tractable network (see Figure~\ref{fig: real_data_illustration}).

We adopted a 4-head Transformer encoder \citep{vaswani2017attention} as the representational backbone, implemented using the \texttt{TransformerEncoderLayer} module in the PyTorch framework with the latent representational dimension set to 8. To enhance stability and generalization, we employed standard data augmentation techniques \citep{shorten2019survey}, specifically \texttt{RandomResizedCrop} and \texttt{RandomHorizontalFlip}, providing diverse views of each image while largely preserving its semantics. We optimized the model using AdamW with a cosine-annealing learning-rate schedule starting at $10^{-4}$ and trained for 500 epochs. For evaluation, we generated representations on the testing set from the learned GenSDR representational function and compared them to those produced by BENN, using configurations matched as closely as possible to ensure fairness.

The third and fourth panels of Figure~\ref{fig: stl_10} present the UMAP visualizations of the GenSDR representations and the BENN representations.
Notably, the GenSDR representations captured considerable semantic information, exhibiting clearly separated clusters aligned with the image categories. In contrast, the BENN representations displayed more pronounced mixing among visually similar classes such as monkey, cat, and dog, indicating inferior class separation. This visual observation was supported quantitatively by Silhouette scores \citep{rousseeuw1987silhouettes}, which measure clustering quality, with higher values indicating better within-cluster coherence and between-cluster separation. We computed the scores using the ground-truth labels to assess the inherent clusterability in the given representational spaces. GenSDR attained a Silhouette score of $0.144$, nearly twice that of BENN ($0.079$). In addition, the distance correlation between GenSDR representations and the original PCs reached 0.823, signifying substantial information recovery and exceeding the corresponding value for BENN (0.808). The learned representational function utilizing GenSDR is thus recommended as a lightweight \textit{student model} that distills sufficient semantic information from the two pre-trained, large-scale \textit{teacher models}. In summary, these results highlight the strong empirical performance of GenSDR and suggest its suitability for a wide spectrum of real-world representation-learning applications.

\section{Conclusion}\label{sec: conclusion}

We present GenSDR, a method for uncovering latent sufficient structures in nonlinear SDR problems by leveraging modern generative models. Building on a reformulation of the conditional velocity field, we construct an appropriate loss function to learn the sufficient transformation. Notably, under suitable conditions, our method achieves population-level validity, demonstrating its exhaustiveness, and sample-level consistency in terms of conditional distributions. The derivation of these sample-level guarantees crucially relies on the use of generative models, thereby providing a principled justification for the GenSDR framework. Moreover, the developed smoothness characterization of the conditional velocity field may be of independent interest. Through incorporating an ensemble strategy, we further extend GenSDR to accommodate non-Euclidean responses, enabling broader applicability. Extensive numerical experiments underscore the empirical effectiveness of GenSDR and highlight its strong capability to comprehensively identify sufficient structures. Although this paper focuses on the more general case of nonlinear SDR, the theoretical framework we have developed is naturally compatible with linear SDR. In this work, the intrinsic dimensionality is assumed to be known; developing consistent estimators of this quantity within our framework is an important direction for future research.

\bibliographystyle{apalike2}
\bibliography{refs}

\end{document}